\documentclass[10pt,journal,compsoc]{IEEEtran}

\usepackage{bm}
\usepackage{color}

\ifCLASSOPTIONcompsoc
  \usepackage[nocompress]{cite}
\else
  \usepackage{cite}
\fi

\ifCLASSINFOpdf
  \usepackage[pdftex]{graphicx}
\else
\fi
\usepackage{amsmath}
\usepackage{amssymb}
\usepackage{bm}
\usepackage{array}
\usepackage{algorithm}

\usepackage{algorithmic}
\usepackage{array}

\usepackage[normalem]{ulem}

\usepackage{url}
\begin{document}
\title{Facial Expression Retargeting from Human to Avatar Made Easy}
\author{
	Juyong Zhang,~~~
	Keyu Chen,~~~
	Jianmin Zheng
	\IEEEcompsocitemizethanks{\IEEEcompsocthanksitem J. Zhang and K. Chen are with the School of Mathematical Sciences,
		University of Science and Technology of China.
		\IEEEcompsocthanksitem J. Zheng is with the School of Computer Science and Engineering, Nanyang Technological University, Singapore.}
}

\markboth{Submitted to IEEE Transactions on Visualization and Computer Graphics}
{Zhang \MakeLowercase{\textit{et al.}}: Facial Expression Retargeting from Human to Avatar Made Easy}

\IEEEtitleabstractindextext{
\begin{abstract}
Facial expression retargeting from humans to virtual characters is a useful technique in computer graphics and animation. Traditional methods use markers or blendshapes to construct a mapping between the human and avatar faces. However, these approaches require a tedious 3D modeling process, and the performance relies on the modelers' experience.
In this paper, we propose a brand-new solution to this cross-domain expression transfer problem via nonlinear expression embedding and expression domain translation. We first build low-dimensional latent spaces for the human and avatar facial expressions with variational autoencoder. Then we construct correspondences between the two latent spaces guided by geometric and perceptual constraints. Specifically, we design geometric correspondences to reflect geometric matching and utilize a triplet data structure to express users' perceptual preference of avatar expressions.
A user-friendly method is proposed to automatically generate triplets for a system allowing users to easily and efficiently annotate the correspondences.
Using both geometric and perceptual correspondences,
we trained a network for expression domain translation from human to avatar.
Extensive experimental results and user studies demonstrate that even nonprofessional users can apply our method to generate high-quality facial expression retargeting results with less time and effort.
\end{abstract}

}

\maketitle

\IEEEdisplaynontitleabstractindextext
\IEEEpeerreviewmaketitle

\IEEEraisesectionheading{\section{Introduction}\label{sec:introduction}}
\IEEEPARstart{F}{acial} expressions are caused by muscle movements beneath the skin of faces. They convey emotional and nonverbal information. Transferring facial expressions from one subject to another is a long-standing problem in computer animation, which is also known as \emph{expression cloning}~\cite{Noh2001ExpressionC} and \emph{retargeting}~\cite{zell2017facial}. The \emph{retargeting} process typically includes capturing facial performances of a source subject and then transferring the expressions to a target subject.
Though the recent development of 3D facial reconstruction~\cite{Guo2017CNNBasedRD, thies2015realtime, Gecer2019GANFITGA} has achieved affordable high performance in real-time , transferring the captured expressions to digital characters remains challenging. This is because current techniques and tools require much effort and work with professional skills, due to
the stylized characteristics of the target subjects.
The popular \emph{blendshape-based animation}, which uses equivalent blendshapes between the source and target characters, is an example that requires experts to control a complicated face rig model to construct 3D avatar blendshapes~\cite{Lewis2014PTBFM}.

\begin{figure}[ht]
\begin{center}
    \includegraphics[width=1 \linewidth]{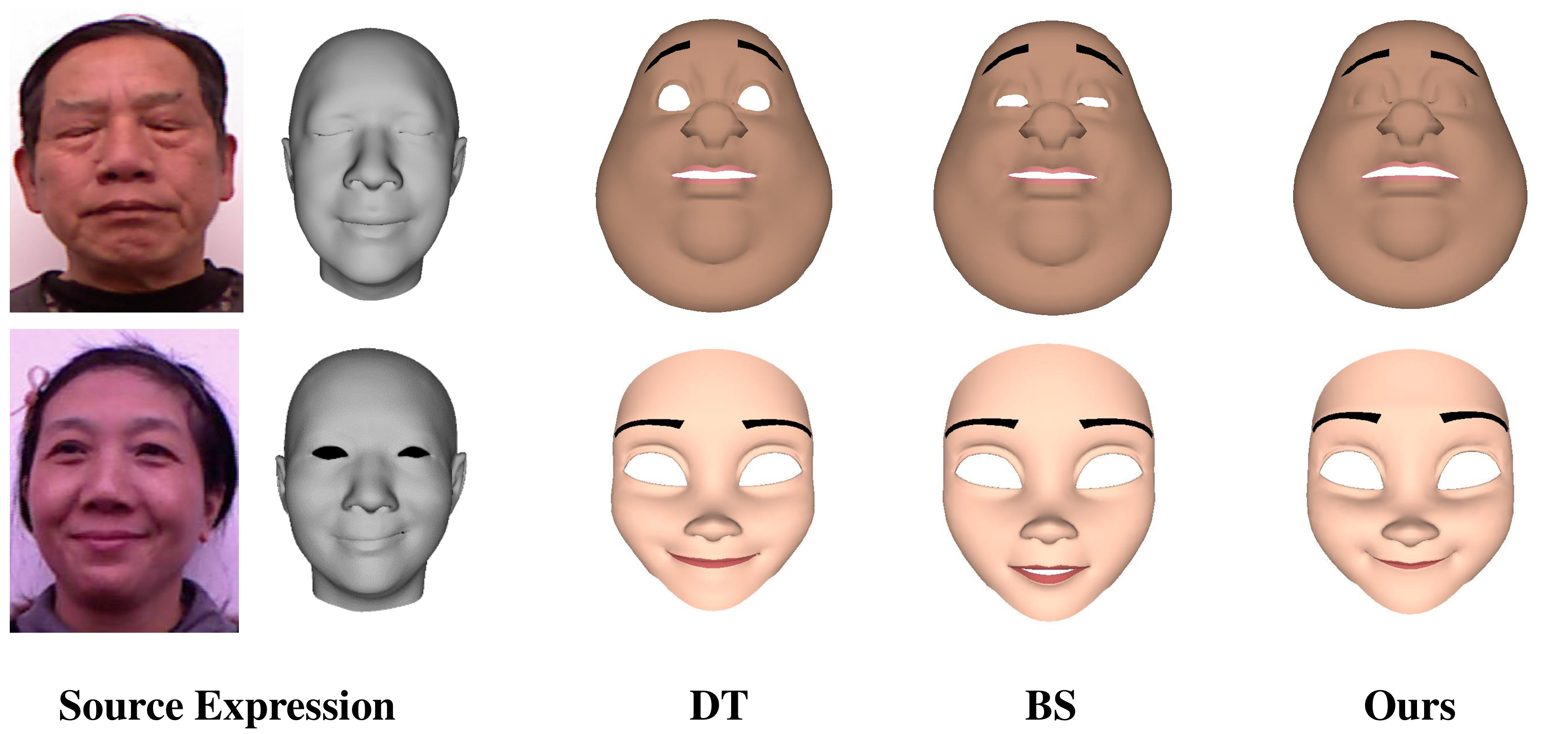}
\end{center}
  \caption{Two input human expressions (picture and mesh) and the retargeting avatar expressions generated by deformation transfer (DT)~\cite{Sumner2004DTT}, the blendshape-based method (BS)~\cite{zell2017facial}, and our method. Our method learns the expression transfer between different characters under geometric and perceptual constraints. It can be seen that the expression semantics are better preserved by our method than the other two methods.\protect\\ \copyright{Face rigs: meryprojet.com, www.highend3d.com}}
\label{fig:pg1example}
\end{figure}

In this paper, we focus on facial expression retargeting that transfers facial expressions from human actors to virtual avatars in a consistent and quality manner (see Fig.~\ref{fig:pg1example}). The consistent manner means that only the expression is transferred; the human identity should not (or less) affect the retargeting results. The quality manner means that the retargeting results should preserve the semantics and fine shape detail of the given human facial expression.
Facial expression retargeting is a cross-domain problem, which remains challenging for existing techniques, such as blendshape-based animation, because avatar faces, in general, are quite different from human actor faces in
terms of shape and expression. Our goal is to develop easy-to-use techniques and tools for this task, which actually involves the following two fundamental problems:
\begin{itemize}
    \item What is a \emph{good} representation of human and avatar facial expressions for retargeting applications?
    \item How can we \emph{easily} construct good correspondences between human and avatar facial expressions?
\end{itemize}

For facial expression representation, a popular model is the blendshape~\cite{Lewis2014PTBFM, FACS}. A linear combination of a set of pre-defined blendshapes defines a low-dimensional parametric space with the coefficients serving as the parameters~\cite{zell2017facial, thies2015realtime, weise2011realtime}. A high-dimensional facial expression can be approximately embedded into the low-dimensional space.
However, if the facial expression is exaggerated, the low-dimensional approximation could be poor. This is because the blendshape-based representation just spans a linear space based on a fixed set of bases, while real facial expressions vary nonlinearly due to complicated emotional movements . Thus some nonlinear methods using deep learning techniques~\cite{tran2018nonlinear, jiang2019disentangled} have been developed.
In this paper, we use a variational autoencoder (VAE) ~\cite{COMA:ECCV18} to embed facial expressions into a latent space. The latent representation of a VAE provides the generalization capability to data unseen from the training set . Furthermore, to make our method independent of (or less dependent on) human identities, we utilize the disentangled 3D face representation learning~\cite{jiang2019disentangled} to extract the expression component from human faces.

Once the expression embedding for human and avatar faces is established, the next question is how to relate the human and avatar expression representations.
A previous approach is {\em parallel parametrization}~\cite{Williams1990PFA, Noh2001ExpressionC}. By building semantically equivalent blendshapes of the source and target, the translation directly copies the blendshape weights from the source parametric space to the target parametric space. However, the construction of semantically equivalent blendshapes is difficult and labor-intensive, requiring
time-consuming modeling process and professional skills.

Our idea is to introduce geometric and perceptual constraints to the process of constructing the correspondence between human and avatar expressions. The geometric constraints are implemented automatically by geometry-based deformation transfer and retrieval. The perceptual constraints are implemented by utilizing the human ability to perceive the subtle changes in facial expressions. Specifically, given a human reference expression, two expressions are automatically selected from the avatar expression dataset to form a triplet. People are asked to choose one of the two avatar expressions: the one which has greater similarity to the human expression. We designed a user-friendly tool for annotation, and the annotators can be nonprofessionals without face modeling experience. The low training requirements for annotators significantly reduces the labor cost and expertise requirement. The annotated results, together with the geometric correspondences, are used to train a domain translation network, which defines a mapping between human and avatar latent spaces of expressions.

In summary, we propose a new facial retargeting framework that comprises variational autoencoder networks to encode human and avatar facial expressions in nonlinear representations, and a domain translation network to translate expression codes across two latent spaces.
In the training stage, we use triplets annotated by nonprofessional people instead of blendshapes created by professional animators to train the semantic correspondences. In the inference stage, our end-to-end system takes as input 3D human face shapes and outputs the retargeting results of avatar expressions. The system can be easily adapted to different virtual characters and can handle various human identities. The main contributions of the paper are twofold:
\begin{itemize}
    \item We propose a novel, two-stage, deep learning based framework for cross-domain 3D facial expression retargeting. The framework is composed of nonlinear facial expression embedding and a domain translation.
    \item We have developed an easy-to-use and robust approach for constructing correspondences between the human and avatar expression latent spaces. The novelty of the method lies in the generation of triplets, the labeling of triplets by nonprofessionals, and the use of labeled triplets for training the domain transfer network.
\end{itemize}

\section{Related Work}\label{sec:related}
\subsection{Blendshape-based Representation}

The blendshape is a prevalent model in the animation and movie industry, and it dates back to the facial action coding system (FACS)~\cite{FACS}. The blendshape model represents complex and varying facial expressions as a linear combination of a group of expressive shapes. It is often used as a basic parametric representation in applications such as facial retargeting~\cite{Hyneman2005HFP, Choe2001Performandriven, Williams1990PFA, Li2010ExamplebasedFR, weise2011realtime}. Constructing a group of blendshapes is, however, labor-intensive and requires modeling experience and skills~\cite{Lewis2014PTBFM}. Recently, 3D facial performance capturing techniques have provided new ways to automatically generate the blendshape models, such as FaceWarehouse~\cite{cao2014facewarehouse} and the 3D morphable model~\cite{blanz1999morphable}. These models also benefit other tasks, for example, face reconstruction~\cite{Guo2017CNNBasedRD, Gecer2019GANFITGA, Thies2018Face2FaceRF}. Despite the existence of several human facial expression blendshapes, there is currently no universal facial expression representation for stylized characters in the digital world, which shares the same semantics . Avatar blendshapes still have to be manually created by artists using professional 3D modeling software like Autodesk MAYA~\cite{maya} and 3DS Max~\cite{harper2012mastering}.

In order to alleviate the expensive cost of constructing avatar blendshapes, methods have been proposed to avoid manual processing. For example, blendshape refinement~\cite{zell2017facial} was proposed to align the motion ranges of human and avatar on manifolds, and a deep learning approach was developed to apply GANs to translate parametric models between subjects~\cite{Ma2018RealTimeFE}. Different from these blendshape-based approaches, our work trains an embedding network with randomly generated data, which represents facial expressions in a nonlinear manner.

\subsection{Mesh Deformation Transfer}
Transferring a deformation from one object to another is a classical geometric processing problem. The concept of deformation transfer is generalized to expression  cloning~\cite{Noh2001ExpressionC}. Given a source face model and a target face model, the expression cloning process transforms the nonrigid deformation (expression) on the source face to the target face. In general, the source and target faces are represented by meshes that may have different connectives . \cite{Noh2001ExpressionC} suggested an approach to establish dense correspondences and transfer the source deformations to the target via per-vertex
displacements. The approach is improved by using deformation gradients~\cite{Sumner2004DTT} and radial basis functions~\cite{Orvalho2008TransferringTR, Song2011CFR, Seol2012SEC}. Other geometric deformation techniques
for facial performance transfer or animation have also been
introduced, such as controllable interactive editing~\cite{Xu2014CHF}, contact-aware transfer~\cite{Saito2013SCF}, and dynamic modeling~\cite{Bouaziz2013OnlineMF, Ichim2015Dynamic3A, Seol2016CreatingAA}.

In example-based animation, traditionally, artists manually transfer the animation of a source character to a target character, which requires great effort~\cite{Bregler2002TurningTT}. Deformation transfer can reduce this effort by transferring a group of key poses to the target automatically, after which the linear combination of blendshapes can help to generate the inbetween  frames~\cite{Lewis2010DMB}. In this process, the mesh deformation transfer requires a set of corresponding points in the source and target models. If the shapes of the source and target models are rather different (for example, in the situation where the source is a human face and the target is an avatar face), the geometrically and semantically consistent correspondences may fail, leading to bad transfer results~\cite{zell2017facial}.

\subsection{Deep Learning for Face Analysis}
Deep learning for face construction and analysis has attracted much attention in recent years. Previously 3D face shapes were often represented by linear models such as blendshapes~\cite{cao2014facewarehouse} and PCA basis~\cite{blanz1999morphable}. With deep learning tools, a nonlinear representation model can be trained for 3D faces by using a variational autoencoder~\cite{kingma2013auto}.
Equipped with the graph convolutional operator~\cite{defferrard2016convolutional} and the mesh deformation representation~\cite{gao2017sparse}, learning-based methods~\cite{COMA:ECCV18, tran2018nonlinear, jiang2019disentangled} have been developed to represent the faces, which are shown to be more powerful and robust than the linear methods.
As for expression analysis, methods have been developed to use either classified categories~\cite{Gross2008MultiPIE,BenitezQuiroz2016EmotioNetAA} or labeled triplets~\cite{Meng2019LSTMBasedFP,Vemulapalli2018ACE} to embed emotional facial expressions into a discrete or continuous space. There are, however, only a few works that consider stylized avatars or cartoon faces.
In~\cite{Aneja2016ModelingSC}, a categorical cartoon expression dataset is created by facial expression artists, and the data in the dataset are labeled via Mechanical Turk (MT)~\cite{Buhrmester2011AmazonsMT}. In \cite{Reed2019UserGuidedFA}, an evolutionary algorithm is developed to generate plausible facial expressions for digital characters.

\section{Overview}\label{sec:overview}
\begin{figure}[htb!]
\begin{center}
    \includegraphics[width=1 \linewidth]{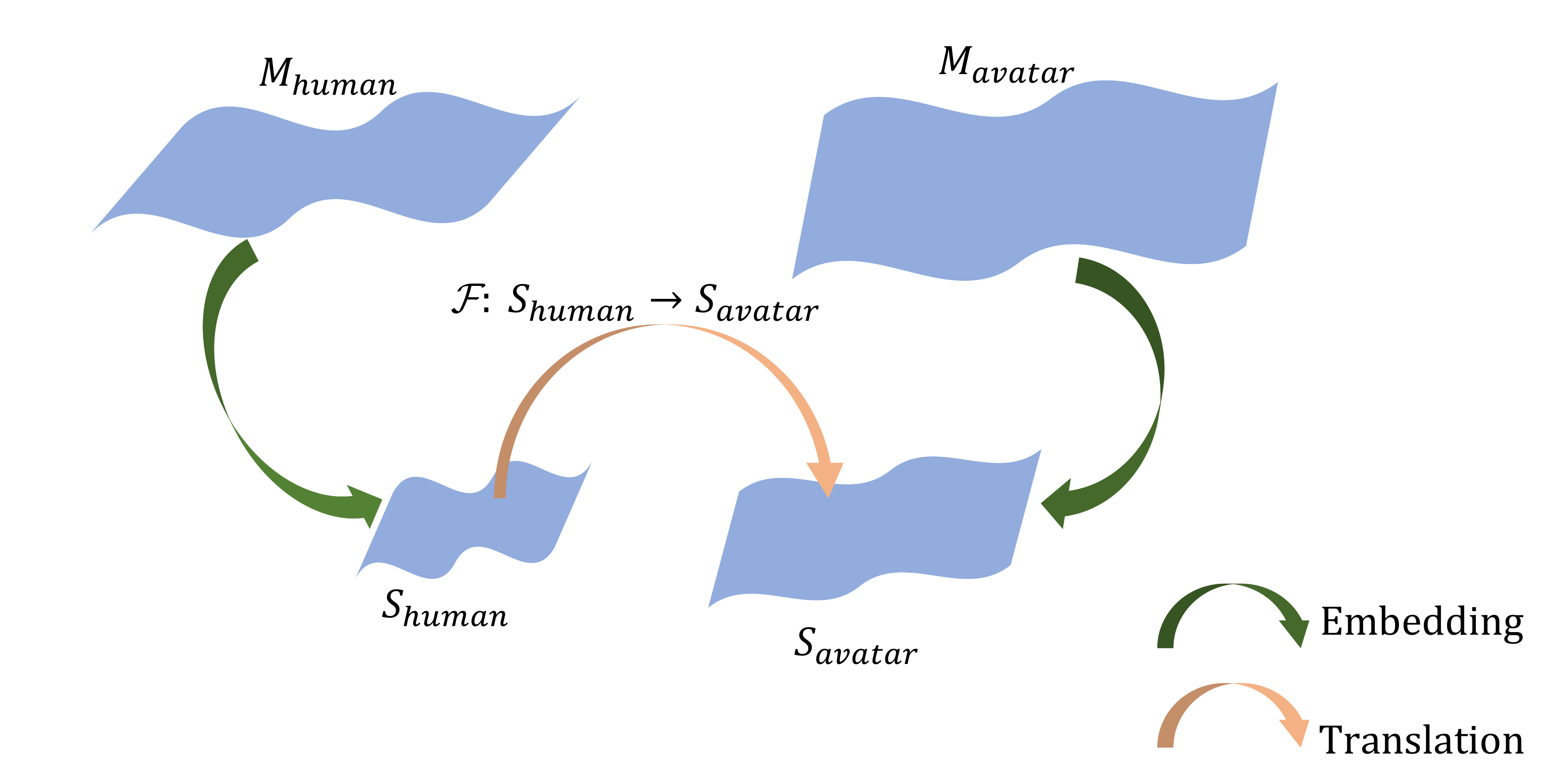}
\end{center}
  \caption{Framework of our facial expression retargeting method. The green arrows represent the facial expression embedding, and the orange arrow represents the domain translation between the two latent spaces.}
\label{fig:framework}
\end{figure}

This section introduces the framework of our facial retargeting.
As illustrated in Fig.~\ref{fig:framework}, the framework consists of two processes: (1) expression embedding and (2) correspondence construction and domain translation network training.
Specifically, let $\mathcal{M}_{human}$ and $\mathcal{M}_{avatar}$ represent the high-dimensional spaces consisting of the human expression meshes and the avatar expression meshes, respectively.
Our approach is to map a given human facial expression in $\mathcal{M}_{human}$ to an avatar expression in $\mathcal{M}_{avatar}$.
The first-stage process is to embed spaces $\mathcal{M}_{human}$ and $\mathcal{M}_{avatar}$ of the human and avatar expressions into low-dimensional latent spaces denoted by $\mathcal{S}_{human}$ and $\mathcal{S}_{avatar}$. The second stage is to establish a cross-domain translation function $\mathcal{F}: \mathcal{S}_{human}\longrightarrow\mathcal{S}_{avatar}$ that relates the two latent spaces. The details of these processes are explained in Section~\ref{sec:embedding} and Section~\ref{sec:transfer}.

\subsection{Facial Expression Embedding}
Note that the blendshape model provides a simple way to embed a dense facial expression model into a low-dimensional parameter space. In fact, a 3D facial expression can be approximately represented as a neutral expression plus a linear combination of a set of blendshapes with the weights $\mathbf{w}=(\mathbf{w}_1, \mathbf{w}_2,\ldots,\mathbf{w}_N)^T \in \mathbb{R}^N$. Denote the neutral expression by $\mathbf{b}^0$ and the blendshapes by $\{\mathbf{b}^i\}_{i=1}^N$. Both $\mathbf{b}^0$ and $\mathbf{b}^i$ are matrices in $\mathbb{R}^{3\times K}$, composed of mesh vertices, where $K$ is the number of vertices. The weights $\mathbf{w}^\star$ for an expression model $\mathbf{m}$ are the solution to the following minimization problem:
\begin{equation}
    \mathbf{w}^\star = \mathop{\arg\min}_{\mathbf{w}\in \mathbb{R}^N} \|\mathbf{m}-(\mathbf{b}^0 + \sum_{i = 1}^N \mathbf{w}_i(\mathbf{b}^i-\mathbf{b}^0))\|_2^2 + E_{reg}(\mathbf{w}),
    \label{blendshapeapproxiamation}
\end{equation}
where $\|\cdot\|_2$ is the $L_2$ norm. The regularization term $E_{reg}$ is introduced to avoid generating large weights, and thus artifacts, when the blendshapes are not orthogonal~\cite{Lewis2014PTBFM}.

However, such an approach has two drawbacks. First, the blendshape-based representation has difficulty in representing expressions outside of the linear span of $\{\mathbf{b}^i\}_{i=0}^N$. If some exaggerated or unseen expressions are omitted during the blendshape construction, the blendshape model may not be able to represent them well. Second, the construction of expression blendshapes is nontrivial. Although there are works~\cite{cao2014facewarehouse,Lewis2010DMB} that try to automate the blendshape construction by human facial performance capturing, creating the blendshapes for avatars still requires a tedious and time-consuming modeling process. Currently, this is usually done by experienced artists who manipulate digital character models with professional software such as Autodesk MAYA.

Inspired by the success of recent data-driven shape analysis methods~\cite{Tan2018VariationalAF,gao2018autounpair}, we propose to construct a VAE for deformable facial expressions; using the VAE enhances the expression representation from linear to nonlinear. In particular, for each avatar character, we train a VAE network with randomly generated expressions and obtain the latent space for the avatar expression.
The latent space of the VAE network can cover most valid expressions of the avatar. For human faces, we use disentangled 3D face representation learning~\cite{jiang2019disentangled} to decompose the faces into identity and expression components. In this way, the learned human expression representation is not affected by human identity differences.

\subsection{Expression Domain Translation}
After obtaining the embedding of human and avatar expressions, our next step is to construct a mapping function $\mathcal{F}: \mathcal{S}_{human}\longrightarrow\mathcal{S}_{avatar}$. In order to faithfully transfer a human actor's performance to a virtual character, two important constraints should be considered:
\begin{itemize}
    \item Geometric consistency constraint: The original and retargeted expressions should have  similar local geometric details.
    \item Perceptual consistency constraint: The original and retargeted expressions should have similar semantics according to human visual perception.
\end{itemize}

\begin{figure}[htb!]
\begin{center}
    \includegraphics[width=1 \linewidth]{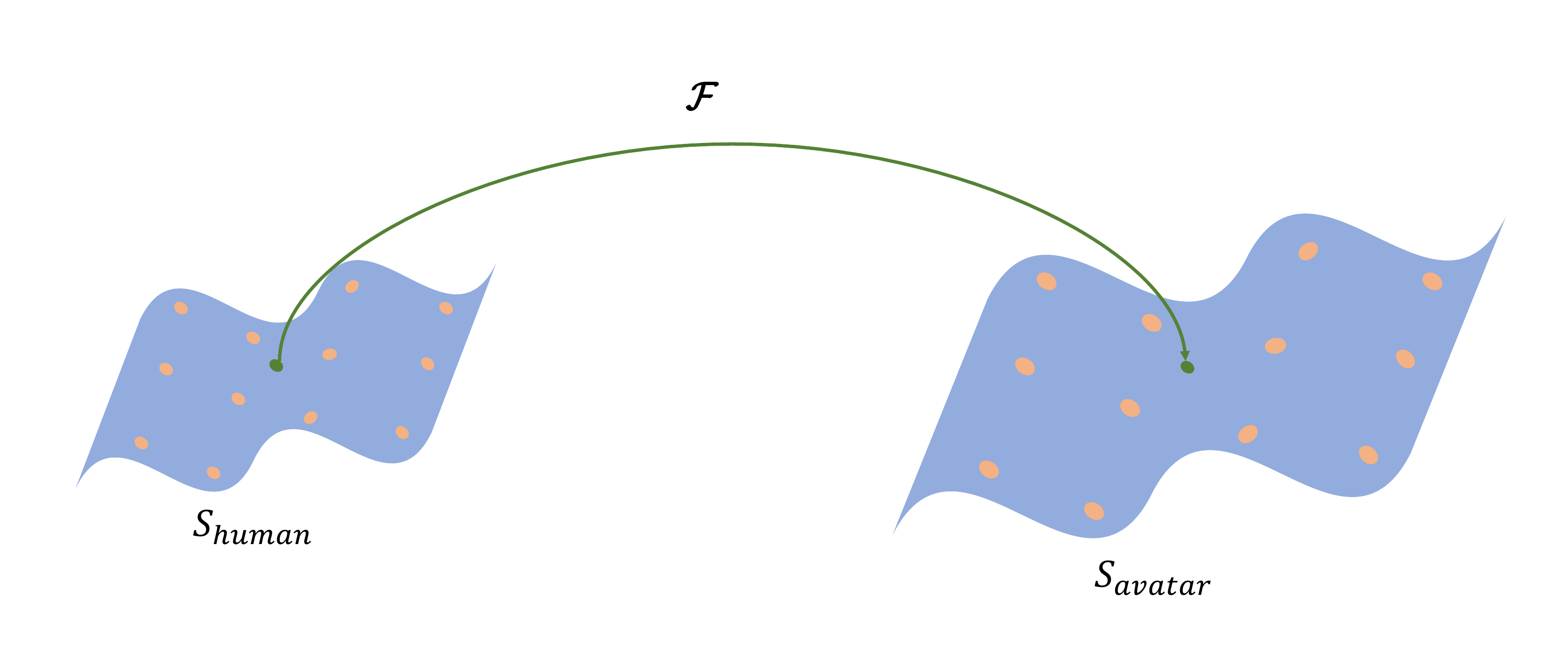}
\end{center}
  \caption{Illustration of typical blendshape-based expression transfer. The semantic equivalent blendshapes induce parallel parametric representations of human and avatar characters. The expression transfer can be easily done by directly copying the blendshape weights from the human parameter space to the avatar parametric space.}
\label{fig:parallel}
\end{figure}

One approach is to create two groups of parallel blendshapes as correspondences between the source and target characters, which is typically used in character animation. With the underlying semantical equivalences, the blendshape weights are directly copied from the source parametric space to the target parametric space, as illustrated in Fig.~\ref{fig:parallel}. However, creating avatar blendshapes corresponding to human references is labor-intensive and requires high skills even for professional animators. Moreover, the quality of the retargeting results is subject to the animators' aesthetic taste.

\begin{figure}[htb!]
\begin{center}
    \includegraphics[width=1 \linewidth]{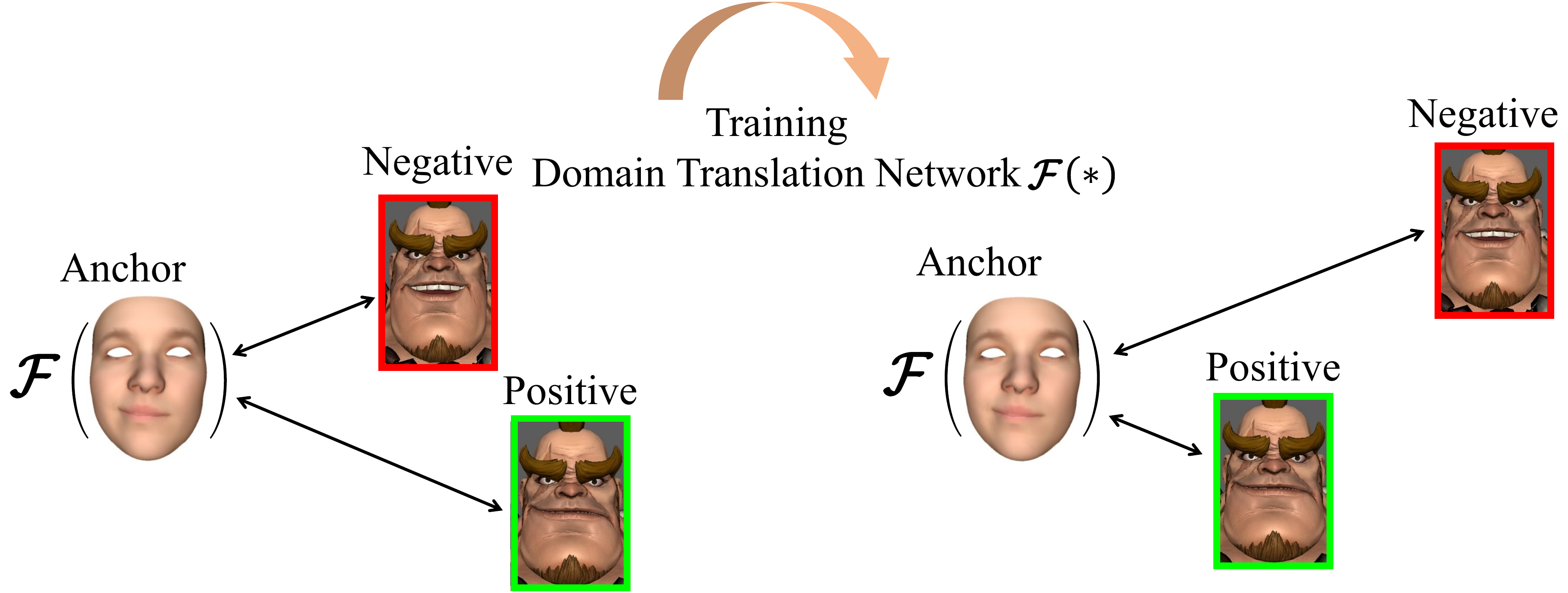}
\end{center}
  \caption{After training the domain translation network using the triplet loss, the anchor point (human expression) is mapped to a location close to the positive point (similar avatar expression) and away from the negative point (dissimilar avatar expression).}
\label{fig:triplet}
\end{figure}

We aim to overcome these issues in constructing the correspondences and present techniques that allow nonprofessionals to engage in the process. For this purpose, we propose to construct correspondences between two expression domains using a triplet data structure, which simply utilizes the common human perception of facial expressions. Consequently, we train a domain translation network~\cite{Schroff2015FaceNet, Weinberger2005DistanceML} using user-annotated triplets (See Fig.~\ref{fig:triplet}) instead of semantic-aware blendshapes.

Given one human expression and two avatar expressions, which together form a triplet, we ask annotators to choose which of the two avatar expressions is more similar to the human expression. Then we utilize the choice to train the cross-domain translation network. Note that all the annotators are nonprofessionals, and the annotation tasks can be simultaneously distributed to them. In this way, we can avoid requiring professional skills and intensive efforts in creating corresponding avatar blendshapes.

To construct the training data, we propose a coarse-to-fine scheme to generate correspondences under both geometric and perceptual constraints. First, we apply the deformation transfer method~\cite{Sumner2004DTT} to compute an initial deformed avatar expression mesh. Second, we automatically retrieve many avatar expressions from the dataset , which are geometrically similar to the initial one. Last, we construct triplets from the retrieved avatar expressions and generate the perceptual correspondences based on user annotation. To accelerate the manual annotation process, we further design a tournament-based strategy to infer new triplet labeling from the annotated ones.

\begin{figure*}[ht!]
\begin{center}
  \includegraphics[width=1\linewidth]{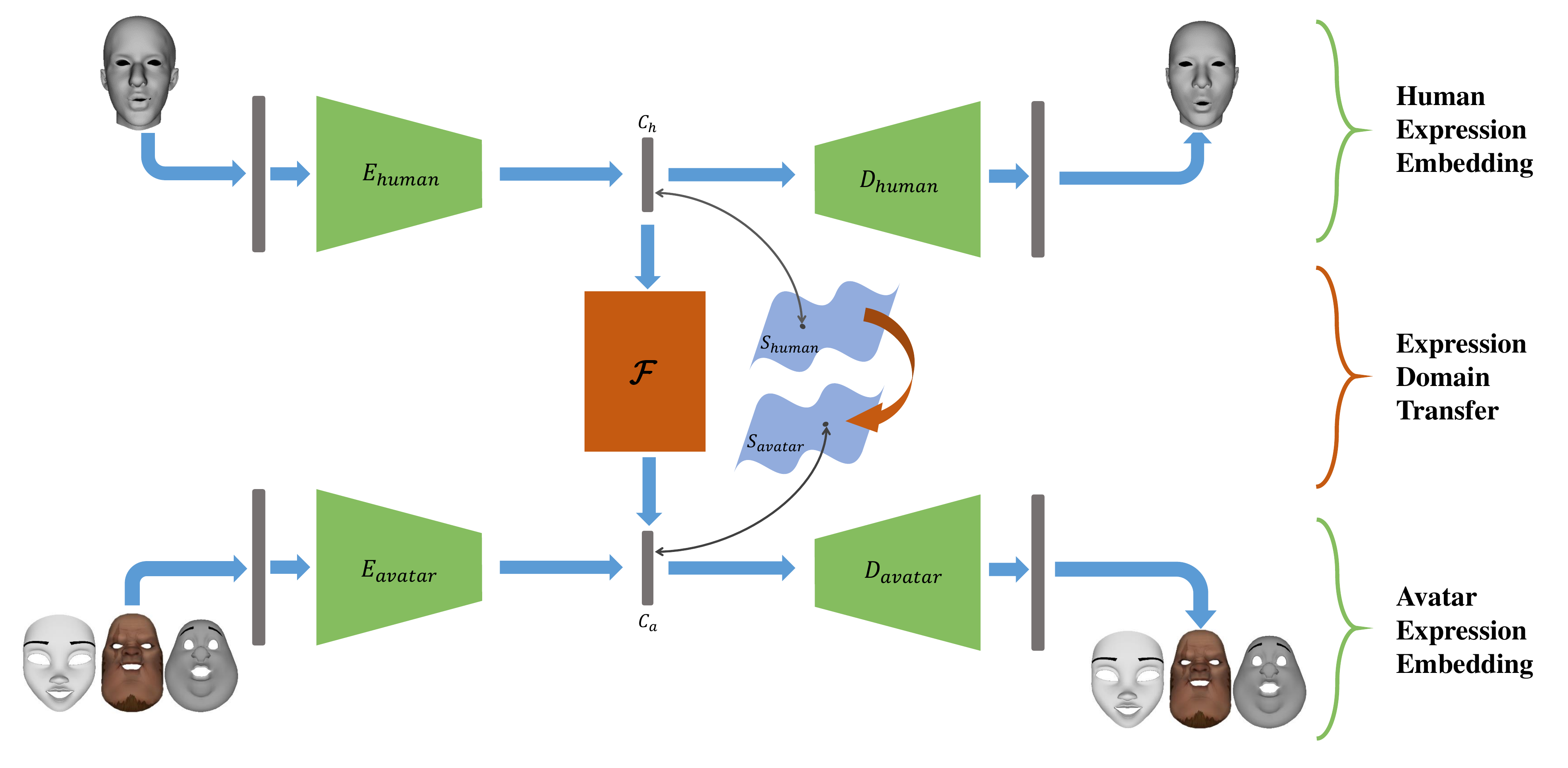}
\end{center}
  \caption{The pipeline of our proposed algorithm. The network includes facial expression embedding and expression domain translation. For humans, we train only one VAE network to embed expressions from different people into an identity-invariant latent space. For avatars, we train an individual VAE network for each character. For human and avatar expression representations, a domain translation network is trained under both geometric and perceptual constraints.\protect\\ \textcopyright{Face rigs: meryprojet.com, Tri Nguyen, www.highend3d.com}}
\label{fig:pipeline}
\end{figure*}

Fig.~\ref{fig:pipeline} shows the pipeline of our proposed facial retargeting approach. The top and bottom branches are separate graph convolutional VAE networks for human and avatar expression representation learning. Across the two latent spaces, a fully-connected network $\mathcal{F}$ is trained for domain translation. In the inference stage, our framework takes as input 3D human face shapes, and it outputs retargeted avatar expression shapes.

\section{Nonlinear Expression Embedding} \label{sec:embedding}
A powerful parametric model, which can cover most deformable facial expression shapes, is a crucial component of any retargeting system. To improve the representation ability of linear models, we employ a VAE network~\cite{Tan2018VariationalAF, jiang2019disentangled} to learn a nonlinear latent representation. The generalization property of VAEs enables the embedded latent space to represent a wider range of expressions, which helps to improve the robustness of the facial retargeting process. Our approach is to train an individual VAE network for each avatar character and train a disentangled network for all human actors.

\subsection{Avatar Expression Domain}
\subsubsection{Data Collection} \label{sec:datacollection}
We obtained rig models of three characters: \emph{Mery}, \emph{Chubby}, and \emph{Conan}. Each rig model has many controllers that are associated with specific muscle movements. For example, the eyebrow controller allows right/left movement, and the jaw controller allows up/down movement. We sample the rigging parameters in a feasible range and thus generate random expressions in 3D shapes.

\begin{figure}[ht!]
\begin{center}
    \includegraphics[width=1 \linewidth]{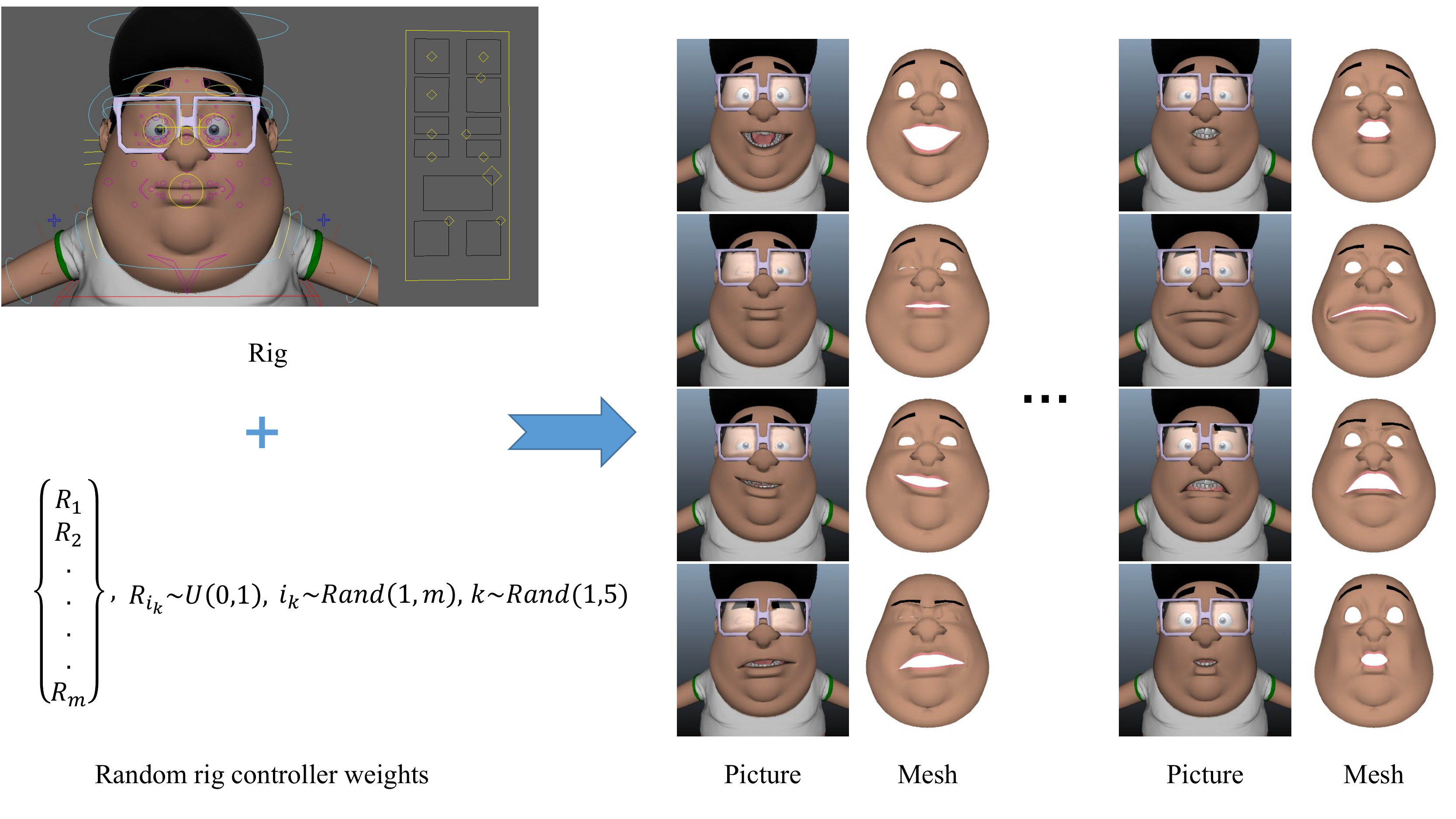}
\end{center}
  \caption{Data collection process. The top-left image shows a facial rig of an avatar character. By randomly sampling valid rig controller parameters, we generate various expressions as shown on the right. $Rand(i,j)$ represents a random sample of integers in interval $[i,j]$, and $U(0,1)$ represents the uniform distribution over $[0,1]$. \protect\\ \copyright{Face rig: Tri Nguyen}}
\label{fig:collection}
\end{figure}

In order to generate reasonable avatar expressions, we limit the number of controller parameters to be sampled. In each round, at most five controllers are chosen with random values of weights falling in $[0,1]$. In this way, we automatically generate 2000 valid expressions for each avatar. Some examples are shown in Fig.~\ref{fig:collection}.

\subsubsection{Deformation Representation Feature}\label{sec:dr}
Given a collection of avatar expression meshes with the same topology, each mesh is converted into a representation using a neutral expression and Deformation Representation (\emph{DR}) feature~\cite{gao2017sparse,wu2018alive, jiang2019disentangled}. Specifically, consider an expression mesh $\mathbf{m}_r=(\mathcal{V}_r, \mathcal{A})$ and denote the neutral expression mesh by $\mathbf{m}_0=(\mathcal{V}_0, \mathcal{A})$, where $\mathcal{V}_r$ and $\mathcal{V}_0$ are the locations of vertices of the meshes and $\mathcal{A}$ is the adjacency matrix encoding the connectivity of the vertices.

The \emph{DR} feature is computed as the local rotation and scale/shear transformations between $\mathbf{m}_r$ and $\mathbf{m}_0$. The computed results are expressed as a 9-dimensional vector for each vertex, and then all these vectors are concatenated into a matrix of $\mathbb{R}^{|\mathcal{V}|\times9}$. Compared to the traditional spatial feature (i.e., cartesian coordinates), the \emph{DR} feature is invariant to global rigid transformations and can better capture the nonrigid deformation of face shapes. For more details of \emph{DR} computation, please refer to~\cite{gao2017sparse}.

\begin{figure*}
\begin{center}
    \includegraphics[width=1 \linewidth]{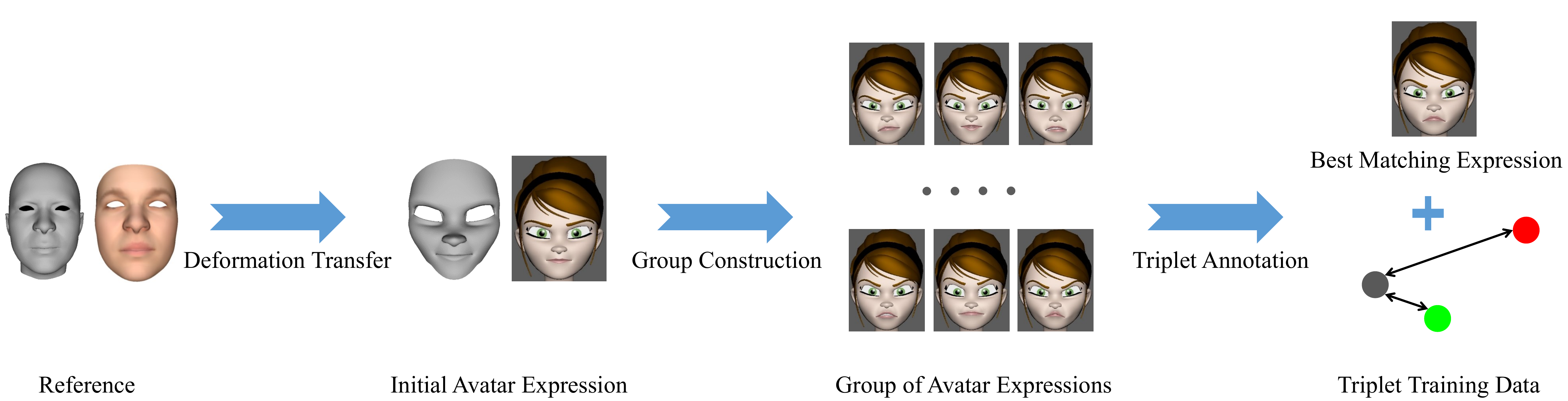}
\end{center}
  \caption{Semantic correspondence construction. Given a human model as the reference, we first generate an initial avatar using \emph{Deformation Transfer}~\cite{Sumner2004DTT}. Then a group of similar avatar expressions is retrieved from the avatar dataset. Finally, multiple users annotate the triplets generated from the groups according to perceptual similarity. The annotation results give the best matching avatar expression as well as the triplet data to train the domain translation network. \protect\\ \copyright{Face rig: meryprojet.com}}
\label{fig:labelling}
\end{figure*}

\subsubsection{Network Architecture and Loss Function}\label{sec:embedding_network}
With the \emph{DR} feature defined on mesh vertices, we use the spectral graph convolutional operator~\cite{defferrard2016convolutional} to process the connectivity information. Because the three avatar characters have different mesh topologies, we train an individual VAE network for each of them. Both the input and output of the VAE network are \emph{DR} features. The input data are embedded into a latent space by multiple fully connected layers with a bottle-neck architecture.

Following the vanilla loss definition in~\cite{Kingma2013AutoEncodingVB}, we train our avatar expression embedding network with two objectives: distribution and reconstruction. That is, the VAE network training seeks to align the latent distribution to data prior and the reconstructed latent code to the original input as much as possible.
Let $\mathcal{G}_{a}$ be the input feature extracted from an avatar expression model, $z$ the latent code, and $D_{avatar}$ the decoder. The loss functions are defined as:
\begin{equation}
\begin{split}
&L_{rec} = \| \mathcal{G}_a-D_{avatar}(z) \|_{1}\\
&L_{kld} = KL(\mathcal{N}(0, 1)\| Q(z|\mathcal{G}_{a})),
\end{split}
\end{equation}
where $\|\cdot\|_1$ is the $L_1$ norm, and $L_{rec}$ and $L_{kld}$ are the reconstruction loss and the Kullback-Leibler (KL) divergence loss, respectively. The KL loss enforces a unit Gaussian prior $\mathcal{N}(0, 1)$ with the zero mean on the distribution of latent vectors $Q(z)$. Once the network is trained, we can take the latent code $z$ as the avatar expression representation in $\mathcal{S}_{avatar}$.

\subsection{Human Expression Domain}
In order to allow different actors as the input to our system, we adopt the disentangled representation learning framework~\cite{jiang2019disentangled} to train a VAE network for human expression embedding. The learned latent space $\mathcal{S}_{human}$ is independent of identities and can easily handle different actors in the inference stage without additional refinement.

Following~\cite{jiang2019disentangled}, we train a VAE network in a disentangled manner to exclude the influence of human identity.
For 3D face models that come from different identities but have the same expressions, we average their shapes and compute the \emph{DR} feature as a ground truth.
The VAE network takes as input the \emph{DR} feature of a face model and outputs a result reflecting the averaging shape. In this way, the identity information is removed from individual data, and the models of different human faces with the same expression are mapped to the same latent code.
The loss functions for training such a human expression embedding network are similar to those for avatars except for the reconstruction target:
\begin{equation}
\begin{split}
&L_{rec} = \| \mathcal{G}_{ref}-D_{human}(z) \|_{1}\\
&L_{kld} = KL(\mathcal{N}(0, 1)\| Q(z|\mathcal{G}_{h})),
\end{split}
\end{equation}
where $D_{human}$ is the decoder, $\mathcal{G}_{h}$ is the original input, and $\mathcal{G}_{ref}$ is the feature of the averaged shape.

\section{Expression Domain Translation} \label{sec:transfer}
This section describes how we construct the correspondences for the facial retargeting task and how we train the translation network to relate the human and avatar expression domains. We first introduce our novel correspondence construction and annotation design in Sec.~\ref{sec:geoDTconstruction} and Sec.~\ref{sec:perceptualconstraint}, respectively, which take both geometric and perceptual constraints into consideration, as illustrated in Fig.~\ref{fig:labelling}. Then in Sec.~\ref{sec:progressivetraining}, we propose a progressive training strategy that helps to stabilize the training process.

\subsection{Geometric Consistency Constraint}~\label{sec:geoDTconstruction}
To maintain the similarity of local geometric deformation between the source and target characters, we first use the deformation transfer algorithm~\cite{Sumner2004DTT} to deform the target avatar models according to the source human expressions. This generates an initial correspondence that can be viewed as the point-to-point correspondences between the human and avatar expressions. However, the geometric deformation often does not give good correspondences for stylized characters due to the large shape difference between the human face and the stylized character face. Therefore, we next relax such point-scale correspondences by retrieving other similar shapes from our avatar dataset based on the $\ell_{2}$ distance in the \emph{DR} feature space.

Thus we obtain a group of avatar expressions that are all somehow similar to the source human expression. Due to the wide range of our collected avatar expressions, we can get various expressions that are similar in semantics but different in shape details.
In this way, we create a group-scale correspondence, and the retrieved group contains more candidate models beyond the deformed mesh.

In the experiment, we used 46 human expressions in FaceWarehouse dataset~\cite{cao2014facewarehouse} as human references. After going through the disentangled expression embedding network, they are embedded into the latent space and the latent codes are denoted by $X=\{x_{1}, x_{2},..., x_{46}\}\subseteq \mathcal{S}_{human}$. Corresponding to each code $x_k$, the latent code of the deformed avatar expression is denoted by $y_k^0\in \mathcal{S}_{avatar}$ and the latent codes of the retrieved group of avatar expressions are denoted by $Y_k = \{y_k^1,..., y_k^P\}\subseteq \mathcal{S}_{avatar}$, where $P$ is the number of the avatar models we retrieve from $\mathcal{M}_{avatar}$. The retrieval is done by searching for the $P$ avatar shapes that are nearest to $y_k^0$ in the \emph{DR} space based on the $L_2$ norm.

In the training stage, we let the expression translation network learn both the point-scale and group-scale correspondences. The translation network $\mathcal{F}$ maps the latent code of the human expressions to the latent code of the avatar expressions.

First, we train the translation network $\mathcal{F}$ with the point-to-point correspondences using the following paired loss function:
\begin{equation}
    L_{P} = \frac{1}{K}\sum_{k=1}^{46}\Vert\mathcal{F}(x_k) - y_k^0 \Vert _2.
\end{equation}
The $L_{P}$ term aims to map the code $x_k$ of a human reference expression accurately to the code $y_k^0$ of the deformed avatar expression. 
Second, the point-to-group correspondences are added into the training with the group-wise loss function:
\begin{equation}
    L_{G} = \frac{1}{KP}\sum_{p=1}^{P}\sum_{k=1}^{46}\Vert\mathcal{F}(x_k) - y_k^p\Vert _2.
\end{equation}

Corresponding to the 46 human source expressions, we have 46 groups of avatar expressions. The $L_{G}$ term tries to map each human expression code $x_k$ to a point that is close to every avatar expression code $y_k^p$ in $Y_k$.

\subsection{Perceptual Consistency Constraint}~\label{sec:perceptualconstraint}
Once we have constructed the group-scale correspondences, we can further improve the domain translation network by fine-tuning the correspondences within groups. The idea is to utilize the human perceptual ability to find better avatar expressions in the same group. Specifically, a triplet comprises a human reference expression that is called an \emph{anchor} point, denoted by $x_a$, and two different avatar expressions. The method automatically generates many triplets, and for each triplet, the users are asked to select the avatar expression that is visually more similar to the anchor point. The selected avatar expression is labeled as a \emph{positive} point $y_p$ and the other one is labeled as a \emph{negative} point $y_n$. The annotated triplet $T(x_a,y_p,y_n)$ reflects the users' preference and can be used to construct the triplet loss ~\cite{Schroff2015FaceNet} to train the domain translation network.

The triplet loss is designed to enforce the \emph{anchor} point to be mapped to a point close to the \emph{positive} point and away from the \emph{negative} point in the training stage, as depicted in Fig.~\ref{fig:triplet}. Thus the loss function is formulated as follows:
\begin{equation}
    L_{T} = max\{0, m + D(x_a, y_p) - D(x_a, y_n)\},
\end{equation}
where $m$ is the parameter of the margin distance (empirically set to 0.2 in our work), and
$D(x_a, y_p), D(x_a, y_n)$ represent the distances between the transferred human expression code $\mathcal{F}(x_a)$ and the positive avatar expression code $y_p$, the negative avatar expression code $y_n$, respectively:
\begin{equation}
\begin{split}
&D(x_a, y_p) = \Vert \mathcal{F}(x_a)-y_p \Vert_{2},\\
&D(x_a, y_n) = \Vert \mathcal{F}(x_a)-y_n \Vert_{2}.
\end{split}
\end{equation}
We aim to minimize $D(x_a, y_p)$ and maximize $D(x_a, y_n)$.

\begin{figure}
\begin{center}
    \includegraphics[width=1 \linewidth]{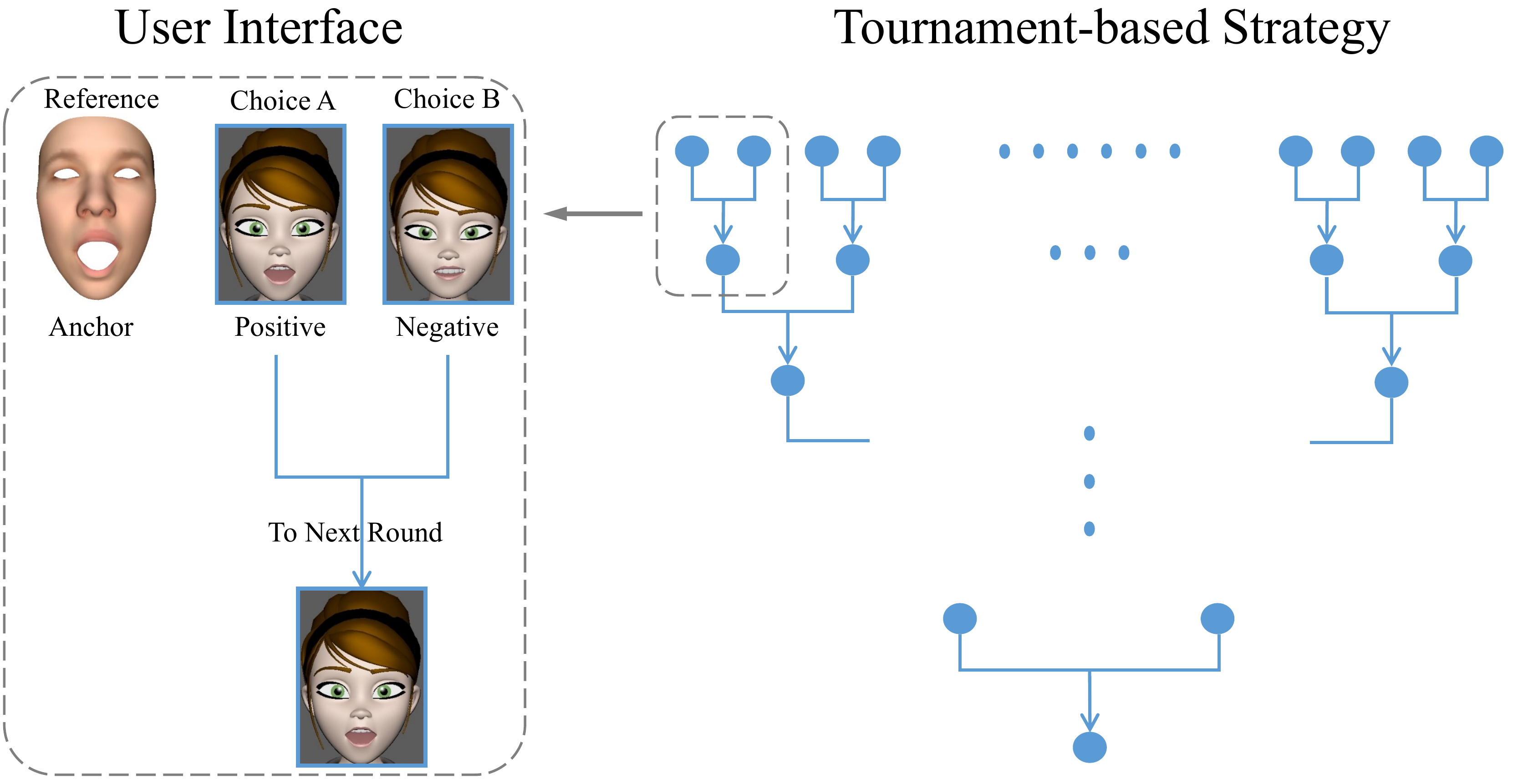}
\end{center}
  \caption{Illustration of our annotation interface and tournament-based triplet generation strategy. In each round, an annotator is shown one human reference expression (i.e., the \emph{anchor}) and two avatar expressions. The annotator determines which avatar is more visually similar to the \emph{anchor}. The selected avatar is labeled as ``positive'' and proceeds to the next round to compete with other \emph{positive} ones. The left avatar is eliminated. The annotation process continues until the final champion (i.e., the best matching avatar expression) emerges.\protect\\ \copyright{Face rig: meryprojet.com}}
\label{fig:interface}
\end{figure}

\begin{figure*}
\begin{center}
  \includegraphics[width=1\linewidth]{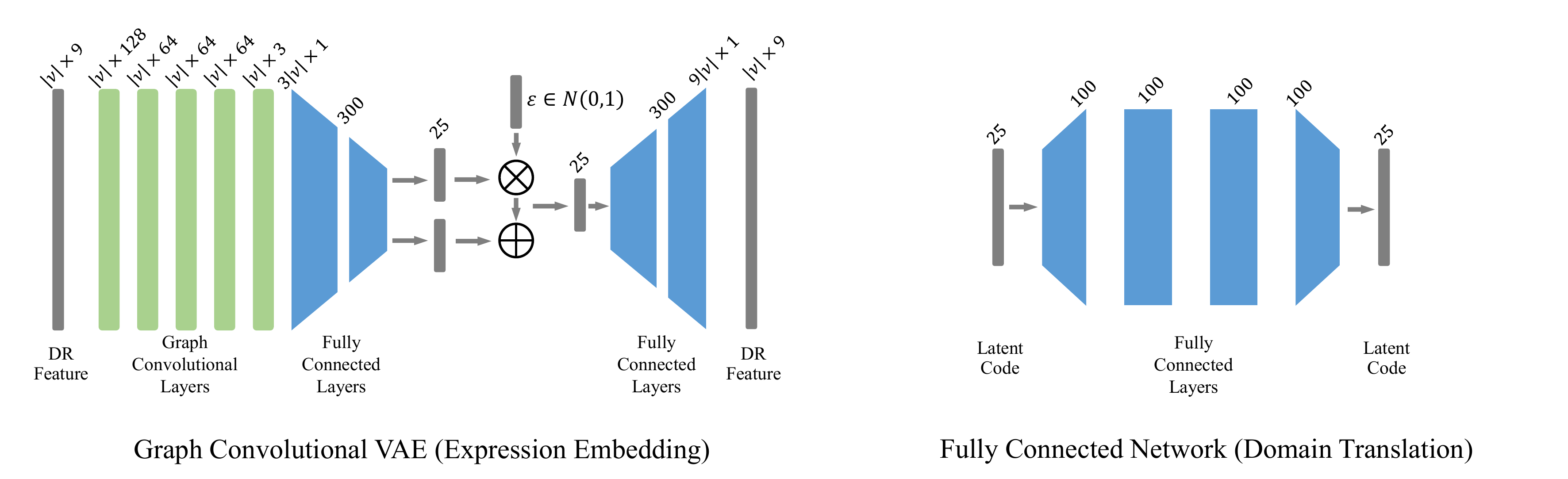}
\end{center}
  \caption{Network structure. On the left is our graph convolutional VAE network which includes graph convolutional layers and fully connected layers. For different topological shapes, the size of input feature will be adjusted. On the right is our domain translation network consisting of FC layers, which relates the two latent spaces.}
\label{fig:networkstructure}
\end{figure*}

To facilitate the annotation process, we build an easy-to-use interface that dynamically generates triplets for multiple users to annotate. We also select triplets that provide useful information or substantially contribute to the correspondence set.
For example, given the same human reference expression $x$, if avatar expression $a$ is judged to be better than $b$ and $b$ is better than $c$, then it is trivial to infer that $a$ should be the \emph{positive} point and $c$ should be the \emph{negative} point in the triplet $T(x, a, c)$.
Therefore we design a tournament-based strategy that enables users to generate such inferences easily, and thus, we accelerate the annotation process.

We are now ready to describe the dynamic triplet generation procedure. As shown in Fig.~\ref{fig:interface}, there are several rounds of elimination in the process. In the first round, the avatar expressions in the same group are randomly paired. For each pair of avatar expressions and the given human reference expression, the annotator chooses which avatar wins. The winner (\emph{positive} one) will proceed to the next round while the loser is eliminated. If an annotator feels that the comparison is too difficult to decide, they can choose the ``draw'' option so that the triplet will be sent to other annotators. The tournament stops when the final ``champion'' is identified. Note that all the annotators use just their own visual perception to make the choices, and they are working in the same system. After the manual annotation process is done, the results are used automatically to augment the training dataset.

In practice, we set the group size to be 16, and we gathered 46 groups in total. With the tournament-based design, we only need to annotate 8+4+2+1=15 triplets for each group and 690 triplets for all 46 groups. We performed this process for three different avatars, so we constructed 2070 triplets. In our experiment, each annotator handled about 700 triplets, and the automatic augmentation generated another 1400 triplets. The annotation process took 10-20 minutes for each annotator and 30-60 minutes in total .

\subsection{Progressive Training}~\label{sec:progressivetraining}
We have introduced geometric and perceptual constraints in building correspondences.
The total loss function can be constructed to include $L_P$, $L_G$, and $L_T$ to train the domain
translation network as follows:
\begin{equation}
    L_{total} = \alpha_P L_{P} + \alpha_G L_{G} + \alpha_T L_{T},
\end{equation}
where  $\alpha_P$, $\alpha_G$, and $\alpha_T$  are trade-off weights balancing the importance of each loss term in different training stages.

Note that three types of correspondences (point-to-point correspondence, group-scale correspondence, and triplet correspondence) are constructed step by step. Their influences on the mapping function decline from the coarse level to the fine level. To achieve better training results, we have to carefully arrange the order of different types of correspondences.

Our progressive network training involves three stages. We set $\alpha_P$ to $1.0$ and $\alpha_G, \alpha_T$ to 1.0e-4  in the first stage to learn the coarse point-scale correspondence. Then we increase the value of $\alpha_G$ to $1.0$ in the second stage to learn the group-scale correspondence. Once the coarse structure of the mapping function $\mathcal{F}$ has been settled, the triplet loss is added to fine-tune the translation network. Therefore, in the third stage, we set $\alpha_T$ to $10.0$ to make the user-guided perceptual correspondence play a major role.

\section{Experiments} \label{sec:experiment}
We first give the implementation details of our neural network structure and training settings in Section~\ref{sec:implementationdetail}. Then in Section~\ref{sec:ablationstudy}, we perform ablation studies to evaluate each component of our method. Next in Section~\ref{sec:comparison} we compare our method with other three competitive approaches: the deep-learning-based method~\cite{gao2018autounpair}, blendshape-based retargeting~\cite{zell2017facial}, and the deformation transfer method~\cite{Sumner2004DTT}. Finally, we discuss two user studies and further experiments to demonstrate the advantages of our method in Section~\ref{sec:userstudy}.

\begin{figure}
\begin{center}
    \includegraphics[width=1 \linewidth]{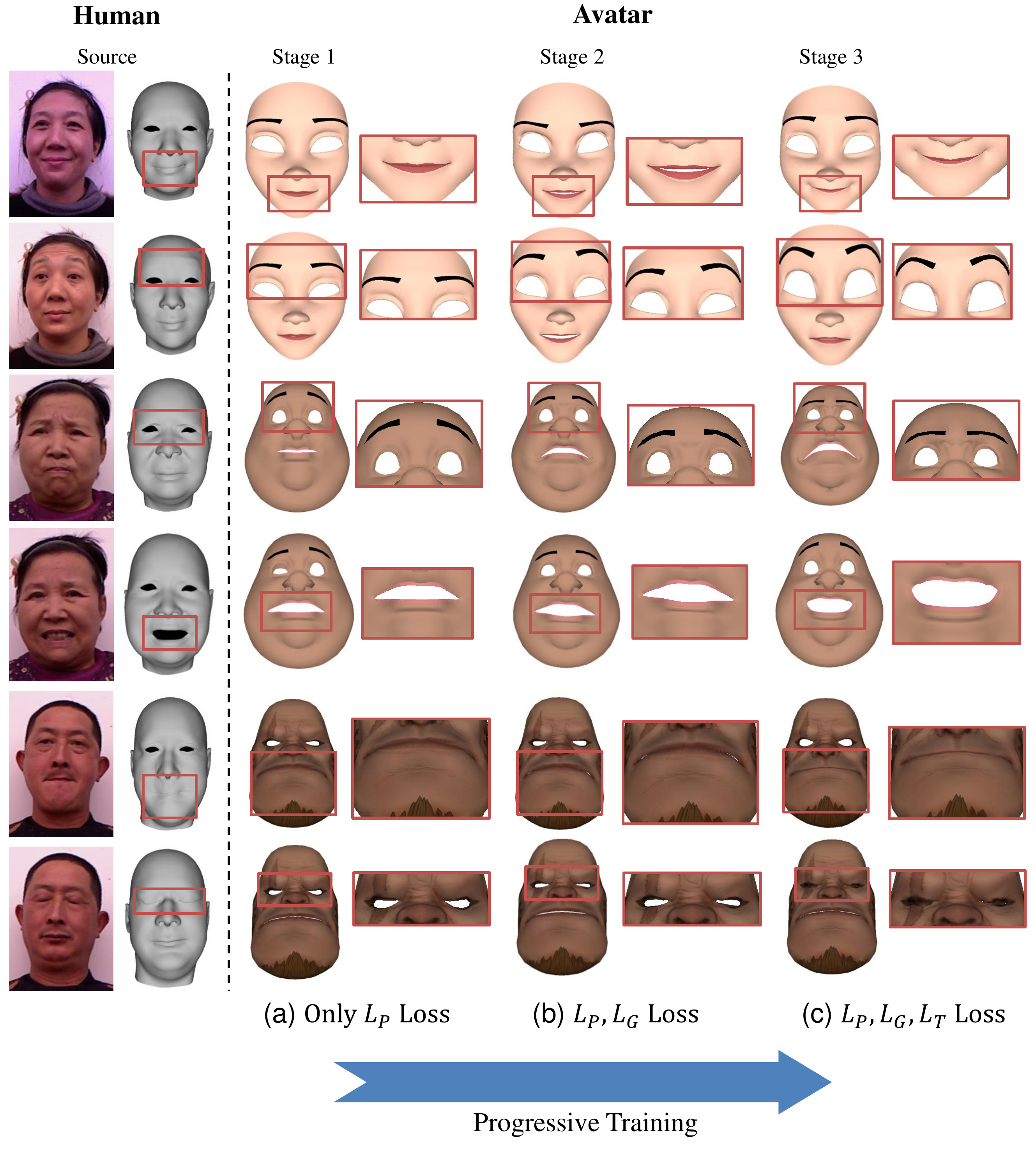}
\end{center}
  \caption{Ablation study on multiple loss functions. On the left are six human expressions. On the right are the training results with different loss functions. It can be observed that by adding $L_G $ and $L_T$ losses, the retargeting results become more natural. \protect\\ \copyright{Face rigs: meryprojet.com, Tri Nguyen, www.highend3d.com}}
\label{fig:ablation}
\end{figure}

\subsection{Implementation Details}~\label{sec:implementationdetail}
Our trained network has two components: facial expression embedding and domain translation. 

For facial expression embedding, two individual VAE networks are trained for the human and avatar, respectively. The human expression embedding network is adapted from \cite{jiang2019disentangled}, in which the disentanglement helps to eliminate the identity variances. We use the same training data as used in \cite{jiang2019disentangled}, which consists of the first 140 subjects and 47 expressions in the FaceWarehouse dataset~\cite{cao2014facewarehouse}. The avatar expression embedding network is a typical graph convolutional VAE network~\cite{COMA:ECCV18}.
We used three different avatar rigs: Mery (\copyright{meryproject.com}), Conan (\copyright{Tri Nguyen}), and Chubby (\copyright{www.highend3d.com}), and trained three individual VAE networks for each of them.
All the VAE networks were trained for 100 epochs with a learning rate of 1e-4  and a decay of 0.6 every ten epochs. We set the batch size to 30 and the dimension of the latent space to 25. The training takes about one to two hours, depending on the number of vertices.

The domain translation network is a 25-25 Multi-Layer Perceptron (MLP). The learning rate decays as 1e-4, 1e-5, and 1e-6  for three different training stages. Fig.~\ref{fig:networkstructure} shows the network structure. For testing, we used the last ten subjects in FaceWarehouse~\cite{cao2014facewarehouse} as source human expressions.

All the training frameworks are implemented in Keras~\cite{chollet2015keras} with Tensorflow~\cite{abadi2016tensorflow} backend. The experiments are run on a PC with an NVIDIA Titan XP and CUDA 8.0. The training code and labeled dataset will be available at \url{https://github.com/kychern/FacialRetargeting}.

\begin{figure}
\begin{center}
    \includegraphics[width=1 \linewidth]{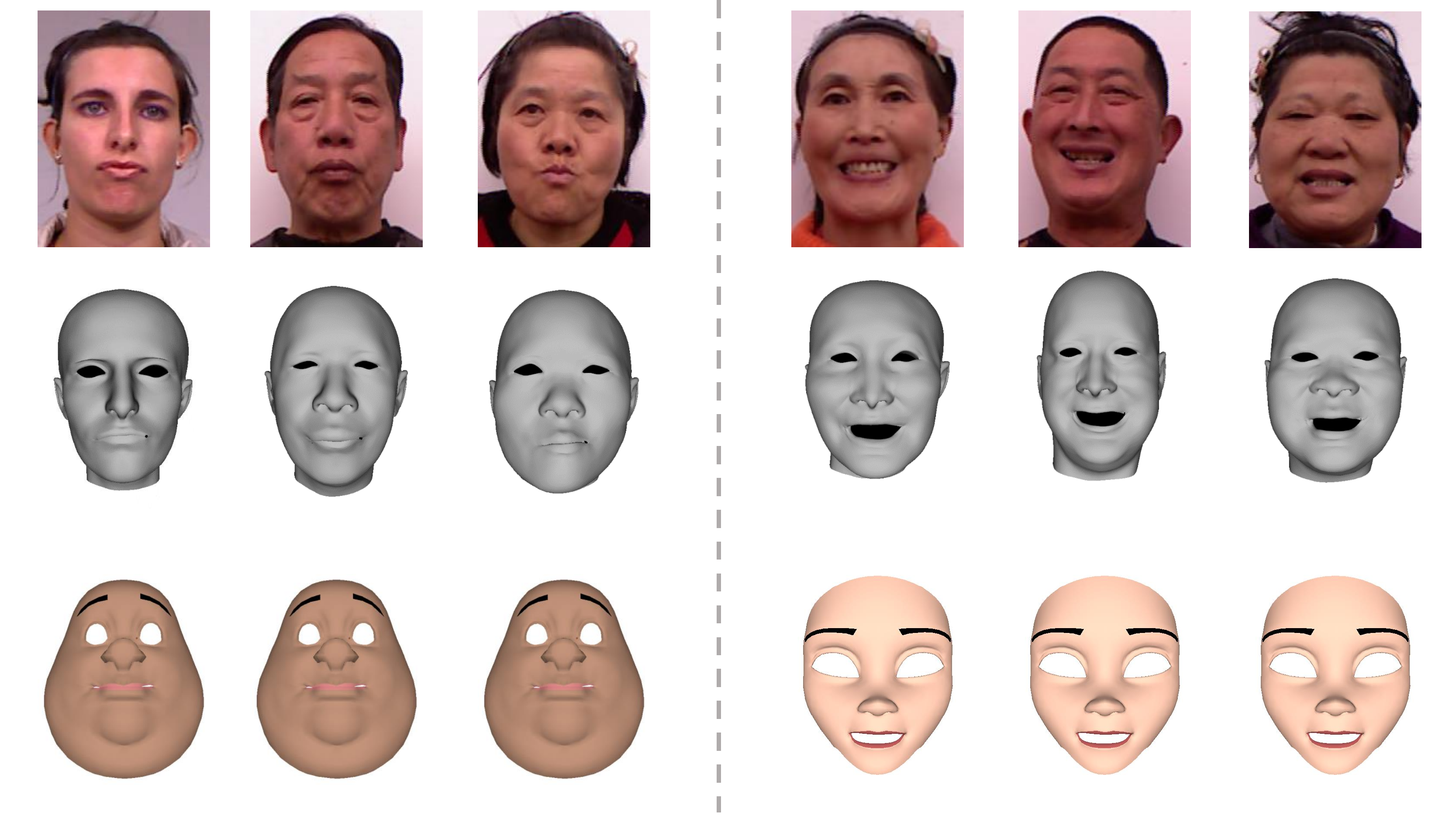}
\end{center}
  \caption{Retargeting results for two expressions, each with three identities. The source human expressions (picture and mesh) and the retargeting avatar expressions are given from top to bottom. The results show that our facial retargeting method is insensitive to identities. \protect\\ \copyright{Face rigs: meryprojet.com, www.highend3d.com}}
\label{fig:stability}
\end{figure}

\subsection{Ablation Study}~\label{sec:ablationstudy}
To evaluate each component of our facial expression retargeting framework, we conducted three ablation studies. The first one was to examine the facial retargeting results at each progressive stage. The second one was to test the sensitivity of our method to the input of different human identities. The last one was to compare our domain translation network with the linear mapping solution~\cite{Lewis2014PTBFM}.

\begin{figure*}
\begin{center}
    \includegraphics[width=0.8\linewidth]{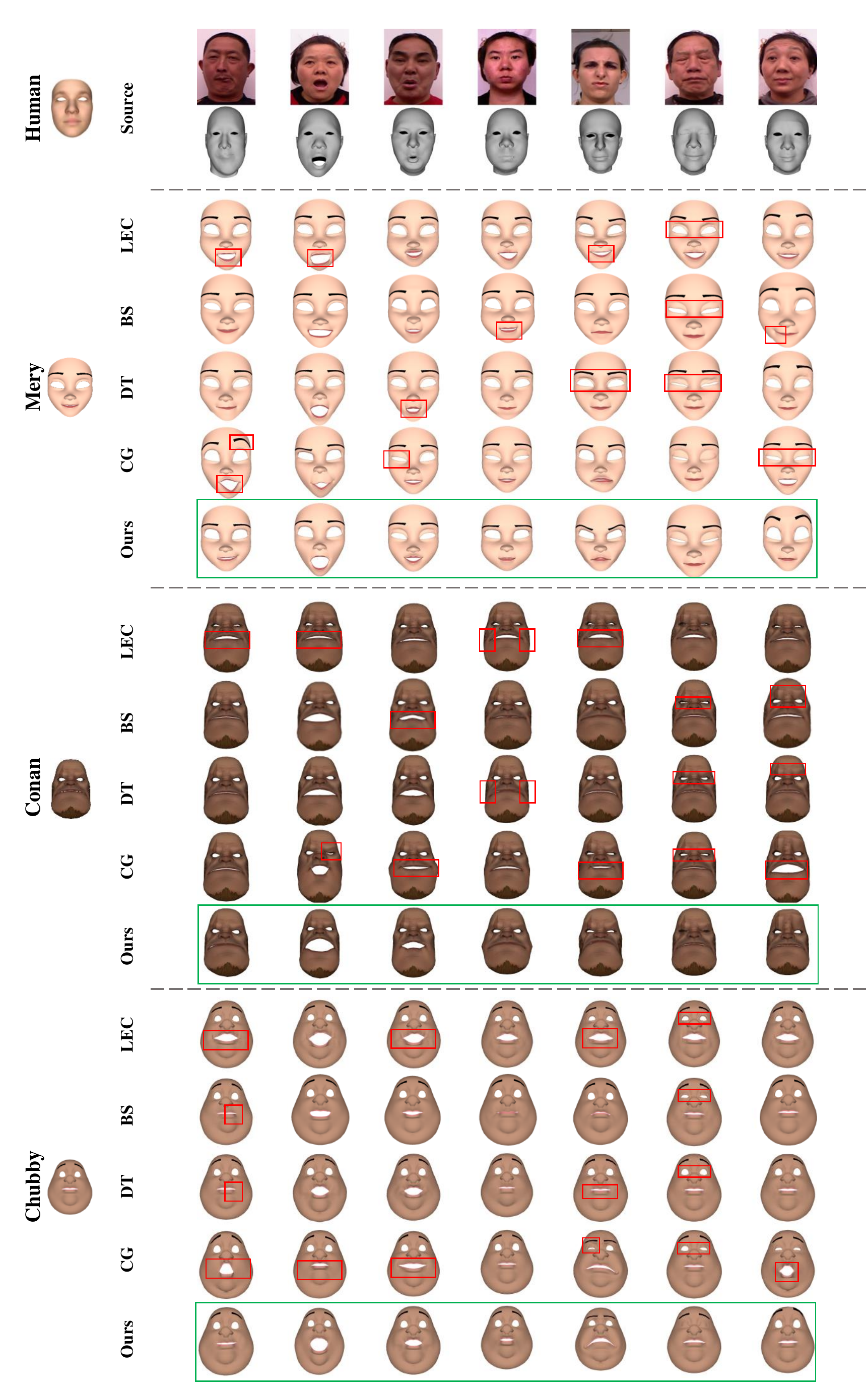}
\end{center}
    \vskip -3mm
  \caption{Comparison of retargeting human expressions to three avatars: \emph{Mery}, \emph{Conan}, and \emph{Chubby}. We show the results generated by linear expression cloning (LEC)~\cite{Lewis2014PTBFM}, the blendshape-based method (BS)~\cite{zell2017facial}, deformation transfer (DT)~\cite{Sumner2004DTT}, CycleGAN+LFD (CG)~\cite{gao2018autounpair}, and our algorithm (Ours). The red rectangles highlight the areas of unfaithfully retargeted results or artifacts. More visualization results can be found in the supplementary video. \protect\\ \copyright{Face rigs: meryprojet.com, Tri Nguyen, www.highend3d.com}}
\label{fig:comparison}
\end{figure*}

\subsubsection{Loss Functions}~\label{sec:lossablation}
To train the translation function $\mathcal{F}$, we propose $L_{P}$ loss for the point-to-point correspondence, $L_{G}$ loss for the group-scale correspondence, and $L_{T}$ loss for the triplet correspondence. Here we perform an ablative study on these three loss functions. Fig.~\ref{fig:ablation} shows the retargeting results generated by the domain translation networks that are trained with (a) $L_{P}$ loss only; (b) $L_{P}$ and $L_{G}$ losses; and (c) $L_{P}$, $L_{G}$, and $L_{T}$ losses. It can be seen that the retargeting results are improved in fine detail by adding the $L_{G}$ and $L_{T}$ loss terms to the network training.

\begin{figure*}
\begin{center}
  \includegraphics[width=1\linewidth]{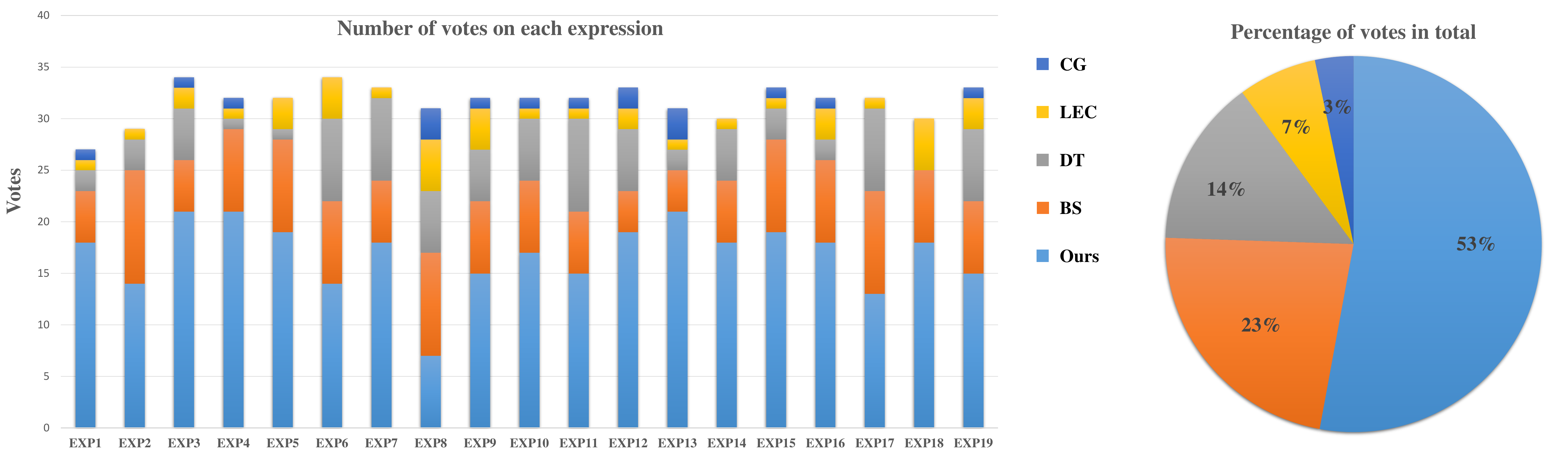}
\end{center}
  \caption{Quantitative results of the user study. The bar graph plots the number of votes that each method receives for each of 19 cases. The pie chart reports the percentage of all the votes each method receives over the total votes. The statistics reveal that both in the single case and in total, our method obtains the highest counts.}
\label{fig:statisticalresult_new}
\end{figure*}

\subsubsection{Identity Independence}
To show the effectiveness of the disentangled human expression representation, we tested the sensitivity of the proposed retargeting method on human identities. We chose subjects with the same facial expression in the testing set of FaceWarehouse dataset~\cite{cao2014facewarehouse} as input. The qualitative results, shown in Fig.~\ref{fig:stability}, indicate that our generated retargeting expressions are independent of human identities, which implies that our method can be applied to different human actors without requiring additional refinement.

\subsubsection{Translation Network}~\label{sec:networkablation}
In~\cite{Lewis2014PTBFM}, the \emph{linear expression cloning (LEC)} approach is proposed to relate the source and target parametric representations without a shared parameterization. Its basic idea is to find a retargeting matrix that specifies the correspondence between the two parameter spaces. In our case, let $\mathbf{s}_k\in \mathbb{R}^{25}$ be the source latent code and $\mathbf{t}_k\in \mathbb{R}^{25}$ be the target latent code. We chose 47 pairs of corresponding expressions, and constructed matrices $\mathbf{S}, \mathbf{T}$ by
collecting $\mathbf{s}_k$ and $\mathbf{t}_k$ as the columns of $\mathbf{S}$ and $\mathbf{T}$ so that $\mathbf{S}, \mathbf{T}\in \mathbb{R}^{47\times 25}$.
Then the expression retargeting matrix $\mathbf{E}\in \mathbb{R}^{47\times 47}$ can be computed as follows:
\begin{equation}
    \begin{split}
        &\mathbf{S}=\mathbf{E}\mathbf{T}\\
        &\mathbf{S}\mathbf{T}^T=\mathbf{E}\mathbf{T}\mathbf{T}^T\\
        &\mathbf{E}=\mathbf{S}\mathbf{T}^T(\mathbf{T}\mathbf{T}^T)^{-1}.
    \end{split}
\end{equation}

To fairly compare our domain translation network with the simple \emph{LEC} approach, we selected 47 human-avatar expression pairs that are best matched according to the user annotation.
The \emph{LEC} approach is regarded as the linear version of our domain translation network. It is capable of capturing the overall structure of the source and target domains, but it does not handle the fine-scale correspondences well. Moreover, if the provided expression pairs are not orthogonal, the linear mapping function may collapse and cause artifacts (see Conan's case in Fig.~\ref{fig:comparison}).

\subsection{Comparison}~\label{sec:comparison}
Now we compare our method with three competitive methods: deformation transfer with CycleGAN~\cite{gao2018autounpair}, blendshape-based animation~\cite{zell2017facial}, and gradient-based deformation transfer~\cite{Sumner2004DTT}.

\subsubsection{CycleGAN+LFD}~\label{sec:cyclegan}
In~\cite{gao2018autounpair}, a deep learning based framework is proposed for shape analysis and deformation transfer. It trains VAE networks for shape embedding and a CycleGAN for latent space translation, which is similar to our approach. Different from our approach that uses carefully designed semantic correspondences of different characters, \cite{gao2018autounpair} adopts the \emph{lighting field distance} (LFD)~\cite{Chen2003OnVS} as the metric to measure the similarities of models from different sets.

In the experiment, we replace our domain translation network with the CycleGAN+LFD network of \cite{gao2018autounpair}. We compute the LFD feature of human and avatar meshes in the same way as ~\cite{gao2018autounpair} and normalize the scores into $[0, 1]$. By comparing the training results of the two methods, it can be observed in Fig.~\ref{fig:comparison} that the CycleGAN+LFD approach often generates undesired results, especially in the areas of eyes and mouth. This is because LFD is a coarse-level feature and incapable of capturing the deformation detail of 3D face shapes.

\subsubsection{Blendshape-based Animation}~\label{sec:blendshapeanimation}
Blendshape-based animation is a common approach in commercial and high-end applications. It is also known as \emph{parallel parameterization}~\cite{Choe2001Performandriven, Hyneman2005HFP, Li2010ExamplebasedFR}. With semantically equivalent blendshapes, the blendshape weights are directly copied from the source parametric space to the target parametric space. However, one problem with this approach is that constructing parallel blendshapes is labor-intensive and requires professional skills. Given the source blendshapes, modelers have to spend much time to create the corresponding target blendshapes. In our approach, we just need triplets, which can be annotated even by nonprofessionals with their visual perception.

For fair comparison in the experiment, we asked an artist to construct a group of avatar blendshapes according to 46 human expressions chosen from the FaceWarehouse dataset~\cite{cao2014facewarehouse}. In \emph{blendshape-based animation}, weight regularization~\cite{Bregler2002TurningTT, Seol2011AFF, Bouaziz2013OnlineMF, zell2017facial} is important to avoid unnatural deformations. We adopted the expression regularization energy term $E_{Retarget}$ from~\cite{zell2017facial} to compute the retargeted blendshape weights. Fig.~\ref{fig:comparison} shows that our method gives better results than the blendshape-based method, especially for the details of eye-closing of Chubby and mouth corner movements of Mery.

\subsubsection{Deformation Transfer}~\label{sec:deformationtransfer}
\emph{Deformation transfer (DT)}~\cite{Sumner2004DTT} uses the deformation gradient to transfer the source deformation to a target. The DT algorithm requires a group of corresponding landmarks labeled on the source and target models. In the experiment, we manually labeled 42 landmarks on both the human and avatar neutral expression meshes. Then we applied the DT algorithm to generate the deformed avatar expressions according to the given human source expressions. Since the DT algorithm is purely based on geometric deformation, the semantics of transferred expressions may be incorrect, especially for stylized characters. As shown in Fig.~\ref{fig:comparison}, some retargeting results generated by the DT approach contain inaccurate details in the large deformation areas like eyelids and mouth.

\subsection{Evaluation}~\label{sec:userstudy}
This section discusses quantitative and qualitative evaluations. We first evaluate the capacity of expression embedding and translation networks. Next, the retargeting results are evaluated based on user preference. We further demonstrate the applicability of our triplet annotation method, also by a user study. Finally, we present animation results, which are driven by sequential video frames.

\subsubsection{Network Capacity}~\label{sec:networkcapacity}
\begin{figure}
\begin{center}
    \includegraphics[width=1 \linewidth]{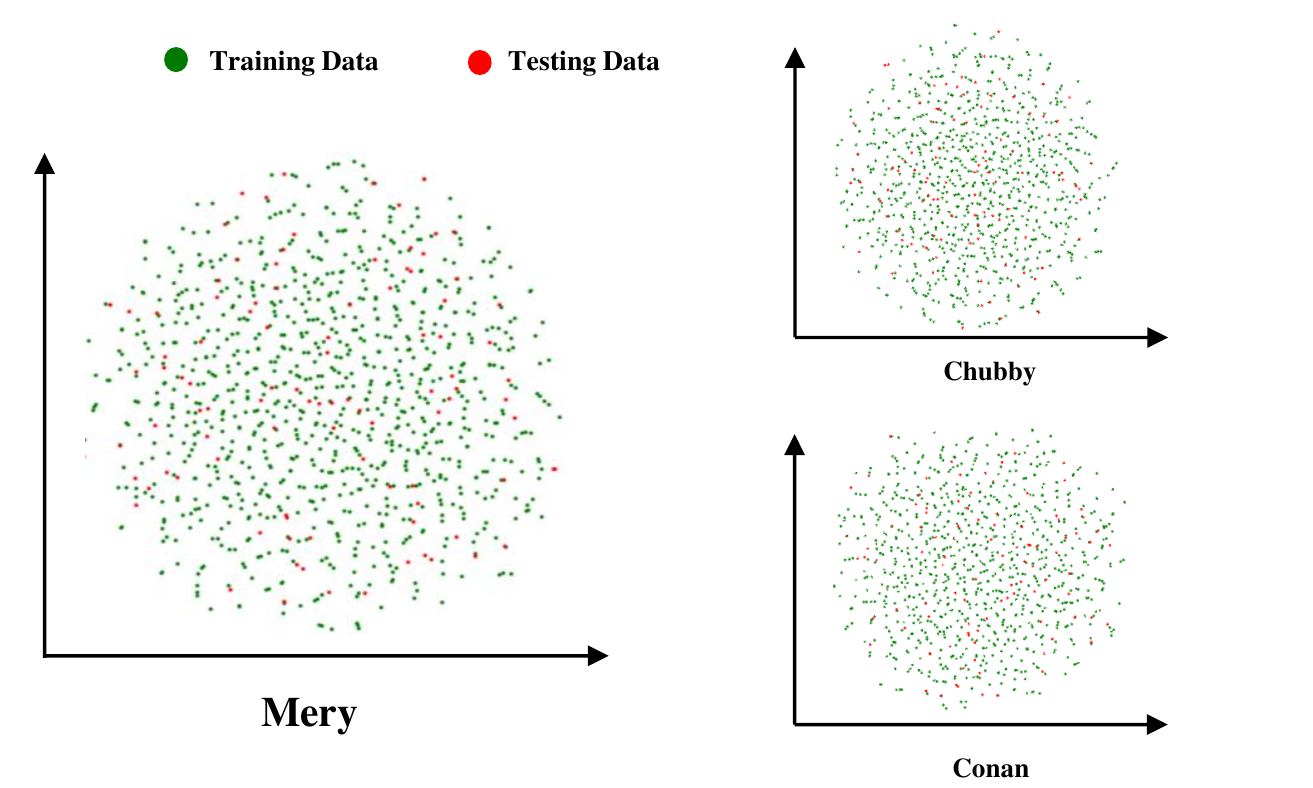}
\end{center}
  \caption{Latent embedding spaces of three VAEs drawn by t-SNE. The green and red points represent training and testing data, respectively. The visualization shows that the learned latent distribution covers the testing data well.}
\label{fig:latentspace}
\end{figure}

To evaluate the VAE networks, we visualize the embedding of training and testing data in Fig.~\ref{fig:latentspace} and calculate the reconstruction errors in Tab.~\ref{tab:reconstructionerror}. From the qualitative and quantitative results, it can be found that the trained model can be generalized to unseen data, and the learned latent space can cover the testing expressions well.

\begin{figure}
\begin{center}
    \includegraphics[width=1 \linewidth]{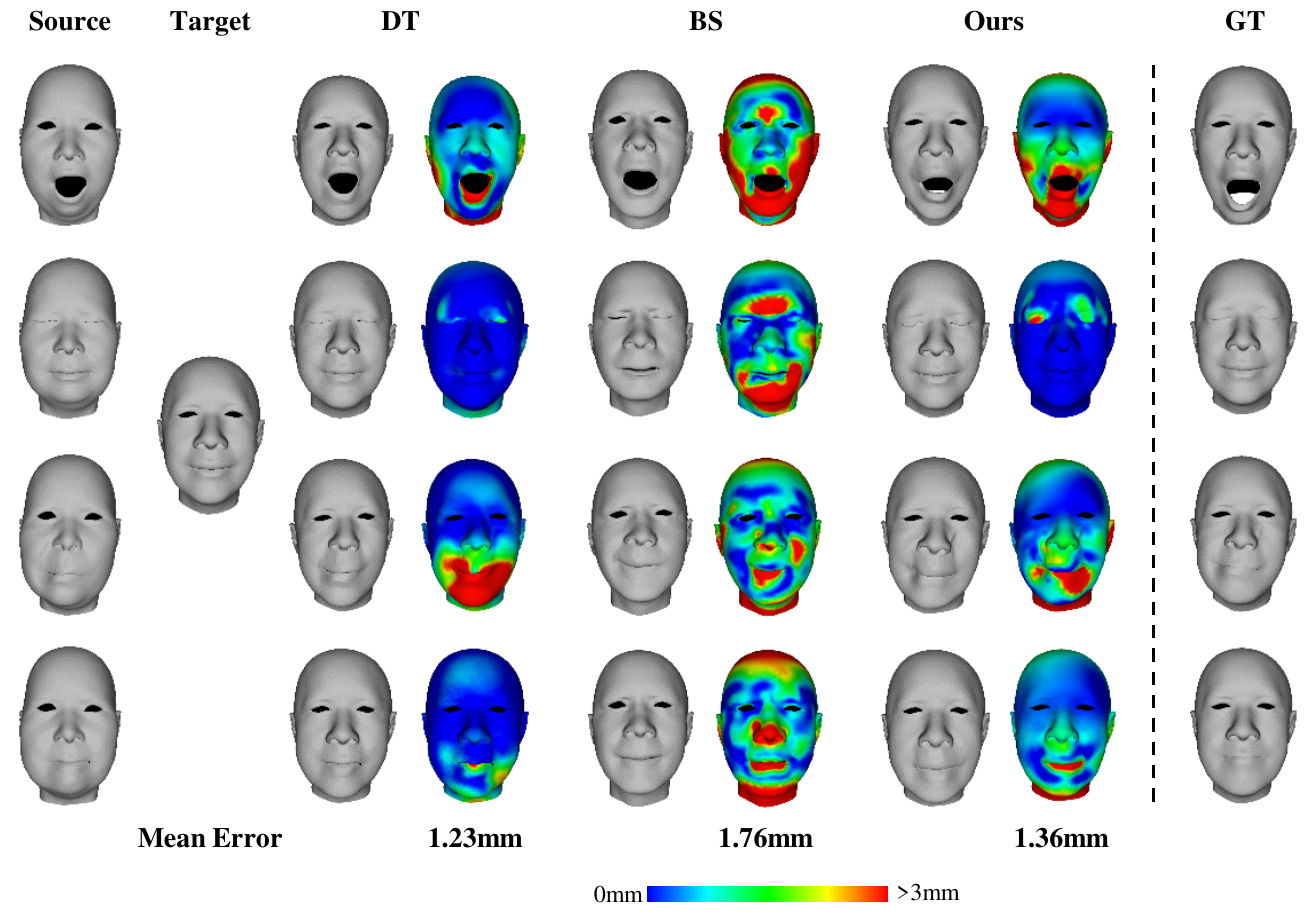}
\end{center}
  \caption{Evaluation of the domain translation network. The first and second columns are the source expressions and the target subject. The right side shows the ground-truth (GT). We compare our method with the deformation transfer method (DT) and the blendshape-based method (BS). The mean errors of vertices between transferred results and the ground-truth are given at the bottom.}
\label{fig:betweenhuman}
\end{figure}

As for the translation network, because there is no ground-truth data (i.e., human-avatar expression pairs), we conducted an alternative experiment. We chose two subjects from the FaceWarehouse dataset~\cite{cao2014facewarehouse} as the source and target characters, and then we trained a translation network following the same settings. We computed the positional errors of vertices between the transferred results and the ground-truth data. The last step was to compare our method with the blendshape-based method (BS) and the deformation transfer method (DT). Both qualitative and quantitative results are given in Fig.~\ref{fig:betweenhuman}. The results show that our method gives smaller errors than BS but larger errors than DT. There are two reasons for DT to have smaller errors than our method in this experiment. First, the source and target human face models share the same topology, so the landmark correspondences are accurate. However, this is generally not true in the scenarios where the source is a human model and the target is an avatar model. Second, the shape difference between human subjects is relatively small, which makes it difficult for our annotators to discriminate the better expression from a triplet.

\begin{table}[!t]
\renewcommand{\arraystretch}{1.3}
\centering
\caption{Reconstruction errors of VAE networks. The second row reports the mean errors of vertices and the third row reports the maximum errors of vertices. All the avatar models are scaled by normalizing the distance between the centers of the two eyes to 10cm.}
\label{tab:reconstructionerror}
\begin{tabular}{|c|c|c|c|}
\hline
Avatar Models & Mery & Chubby  & Conan \\
\hline
Mean Errors & 1.723mm & 1.987mm & 2.125mm\\
\hline
Max Errors & 3.712mm & 4.012mm & 4.673mm\\
\hline
\end{tabular}
\end{table}

\begin{figure}
\begin{center}
    \includegraphics[width=1 \linewidth]{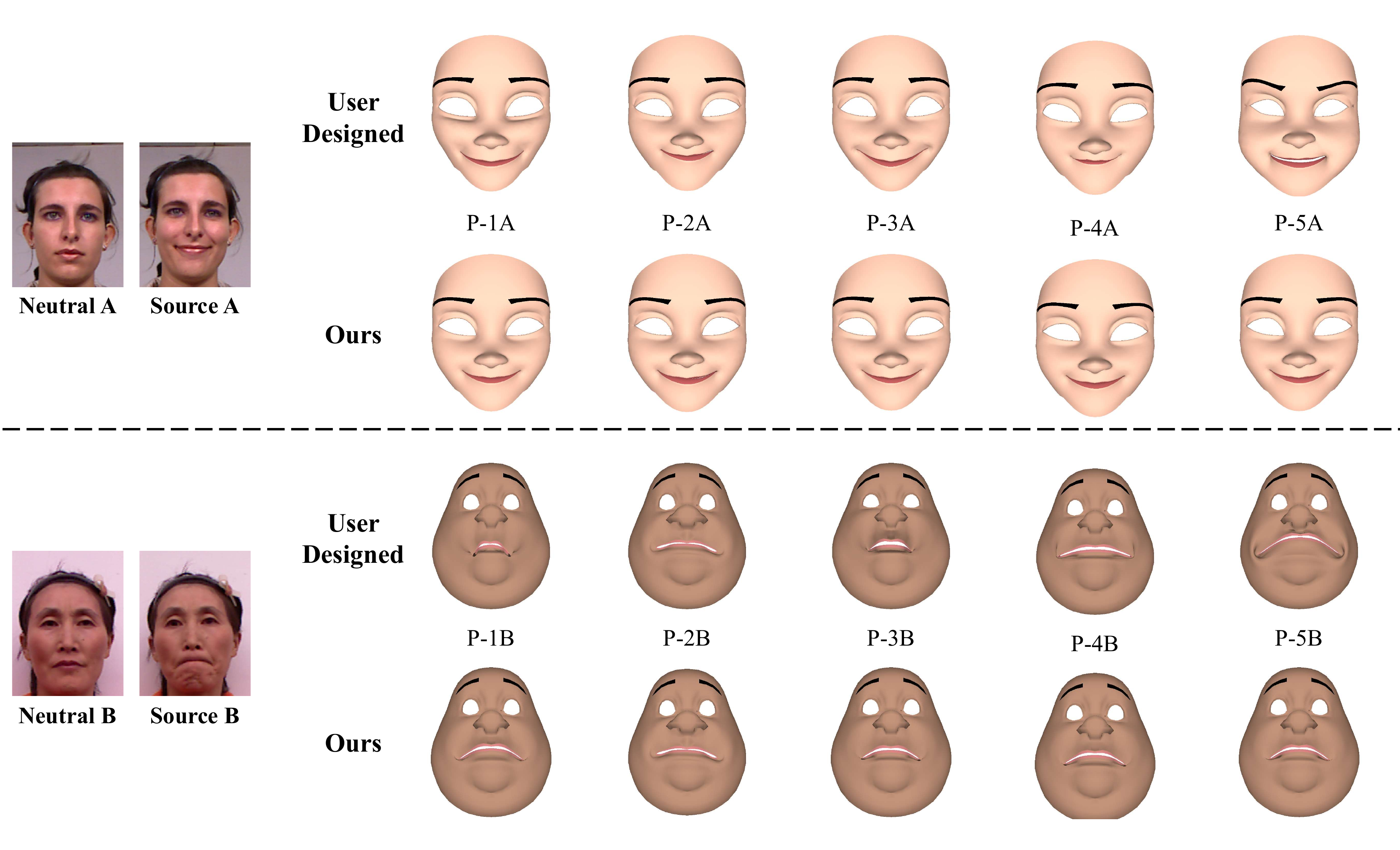}
\end{center}
  \caption{Results created by five participants. ``User-designed'' shows the avatar expression models created by using the Autodesk Maya rigging system. ``Ours'' shows the retargeting results created by using our triplet annotation and network training tool.
  P-$i$A and P-$i$B represent the results created by the $i$-th participant. \protect\\ \copyright{Face rigs: meryprojet.com, www.highend3d.com}}
\label{fig:userstudy}
\end{figure}

\subsubsection{Satisfaction Rate}~\label{sec:satisfaction}
To qualitatively evaluate the facial retargeting results, we invited 24 participants to vote for the five methods listed in Fig.~\ref{fig:comparison}. In each round, five retargeting expression results were shown to them in random order. The participants choose at most two results that they like most. We received 582 votes. Fig.~\ref{fig:statisticalresult_new} depicts the distribution of the votes each method received for 19 different expressions and the overall percentages. Our method received 53\% of votes, and outperforms the others: blendshape-based retargeting~\cite{zell2017facial} (23\%), deformation transfer~\cite{Sumner2004DTT} (14\%), linear expression cloning~\cite{Lewis2014PTBFM} (7\%) and CycleGAN+LFD~\cite{gao2018autounpair} (3\%).

\begin{table}[!t]
\renewcommand{\arraystretch}{1.3}
\centering
\caption{Differences between our method and the blendshape-based method in four aspects.}
\label{tab:applicability}
\begin{tabular}{|c|c|c|}
\hline
Method & Blendshape & Ours\\
\hline
\hline
Time & Three to five hours & Annotation: half to one hour\\
     &  & Training: one hour\\
\hline
Operation & Difficult & Easy\\
\hline
Expertise & Required & Not Required\\
\hline
Stability & Not stable  & Stable \\
 & Artist-specific & User-independent \\
\hline
\end{tabular}
\end{table}

\subsubsection{Applicability}~\label{sec:applicability}
One motivation of our work is to alleviate the intensive labor cost and the professional skills required in semantically equivalent blendshape construction. In this experiment, we conducted a user study to validate the applicability of our method.
Specifically, we invited five participants to try our triplet annotation tool. The annotation results generated by each of them were used to train a domain translation network.
Meanwhile, we taught the participants the basic usage of the Autodesk MAYA rigging system. Then they also used the system to model the avatar expressions corresponding to the given source expressions. They were asked to create only a few avatar blendshapes.

We observe that the participants were more pleased to use our triplet annotation tool than the 3D modeling software. The qualitative results, shown in Fig.~\ref{fig:userstudy}, indicate that for people who do not have much 3D modeling experience, our method can help them to generate more consistent and faithful facial retargeting results. The major differences between our method and the blendshape-based method in four aspects are listed in Tab.~\ref{tab:applicability}. Overall, our method is easy-to-use, time-saving, and user-independent. Moreover, it does not require any professional-level knowledge of 3D expression modeling.

\subsubsection{Video Animation}~\label{sec:video}
\begin{figure}
\begin{center}
    \includegraphics[width=1 \linewidth]{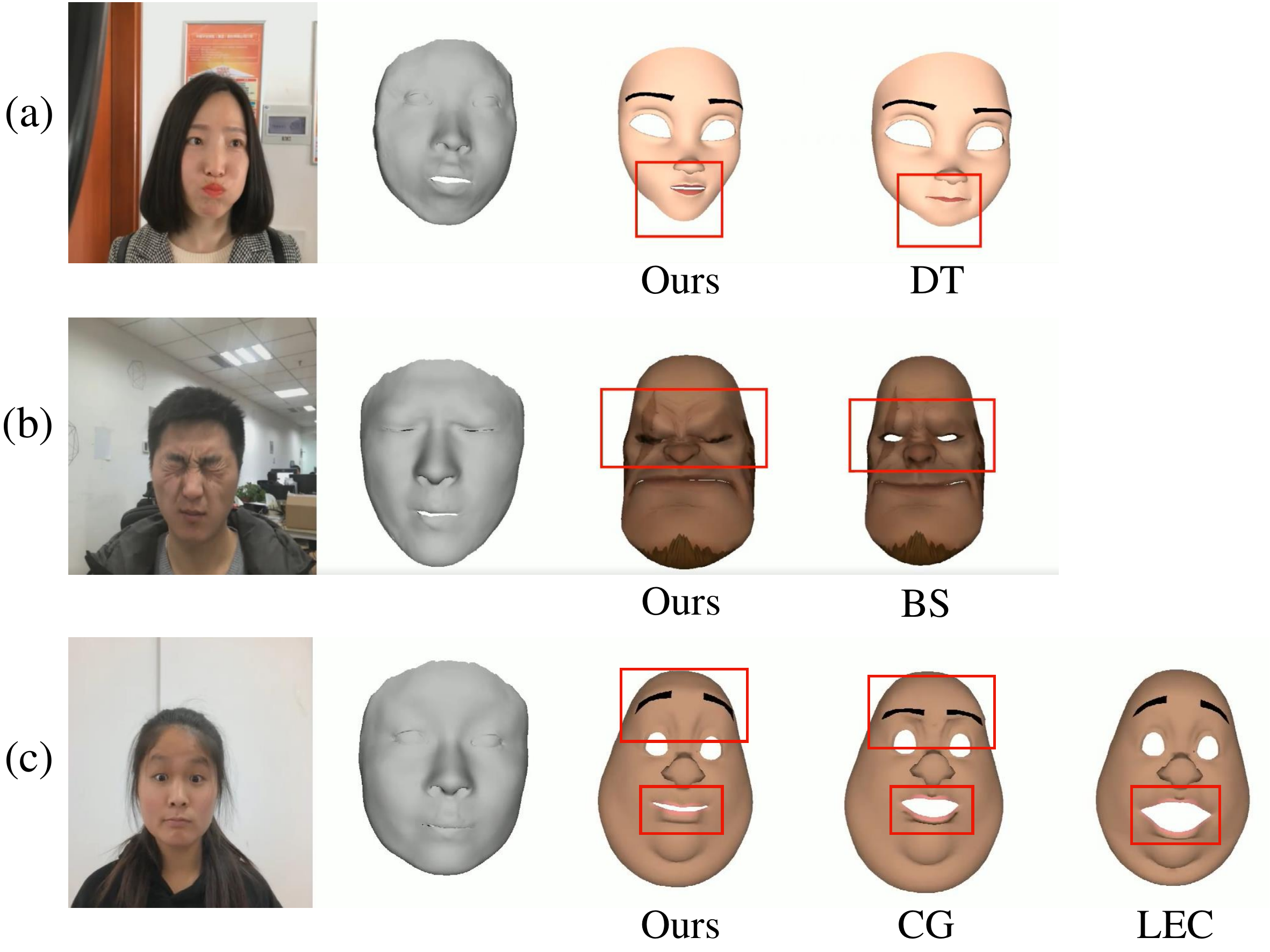}
\end{center}
  \caption{Selected frames from the video animation sequences. (a) Comparison of our method and deformation transfer~\cite{Sumner2004DTT}; (b) comparison of our method and the blendshape-based method~\cite{zell2017facial}; (c) comparison of our method, CycleGAN+LFD~\cite{gao2018autounpair}, and linear expression cloning~\cite{Lewis2014PTBFM}. Please refer to supplementary video for more results.}
\label{fig:videoframe}
\end{figure}

To further demonstrate the performance of our method, we ran our method by taking video frames as input. In general, our method can be integrated into any facial performance capture system, as long as sequential 3D face shapes can be obtained. In this experiment, we employed the 3D face reconstruction method of \cite{Guo2017CNNBasedRD} to reconstruct 3D face shapes from video frames.
Fig.~\ref{fig:videoframe} displays a few selected frames of the output video animation. It can be observed that (a) compared to the deformation transfer method~\cite{Sumner2004DTT}, our method generates more natural results reflecting eyelids and lip motions; (b) compared to the blendshape based method, our method conveys more accurate expressions (for example, closing eyes); and (c) compared to the linear expression cloning and CycleGAN, our method is more stable. The entire animation results can be found in the supplementary video.

\section{Discussion }
While the experiments have shown that our facial expression retargeting method can produce good avatar expressions in an easy-to-use manner, the method also has limitations. First, the annotation and training process is required for each new avatar. That is, for each individual virtual character, we need to train a separate VAE network to embed its expression deformations and manually label the triplet samples for training the domain translation network. Second, the quality of our resulting animation depends on the annotator's carefulness and subjectivity in processing the applications.

In the future, we will explore potential solutions to these issues. For example, in order to avoid re-training VAE models for new avatars, different avatar models could be jointly embedded into a universal latent space by adopting transfer learning methods. Consequently, the domain translation process can be replaced by domain adaptation to embed different character shapes into the same latent expression domain. Also, the efficiency of our triplet annotation might be improved by using active learning~\cite{Settles2012ActiveL}.

\section{Conclusion}
We propose a novel method for facial expression retargeting from humans to avatar characters. The method consists of facial expression embedding and expression domain translation. The two technical components are realized by training two advanced variational autoencoders and a domain translation network.
By employing carefully designed loss functions as well as triplet annotation by nonprofessionals, our method is easy-to-use and does not require professional 3D modeling skills. Both qualitative and quantitative experimental results demonstrated that our method outperforms previous methods.

\section*{Acknowledgments}
This research is partially supported by the National Natural Science Foundation of China (No. 61672481), Youth Innovation Promotion Association CAS (No. 2018495), Zhejiang Lab (NO. 2019NB0AB03), NTU Data Science and Artificial Intelligence Research Center (DSAIR) (No. 04INS000518C130), and the Ministry of Education, Singapore, under its MoE Tier-2 Grant (MoE 2017-T2-1- 076).

\bibliographystyle{IEEEtran}
\bibliography{reference}

\begin{IEEEbiography}[{\includegraphics[width=1in]{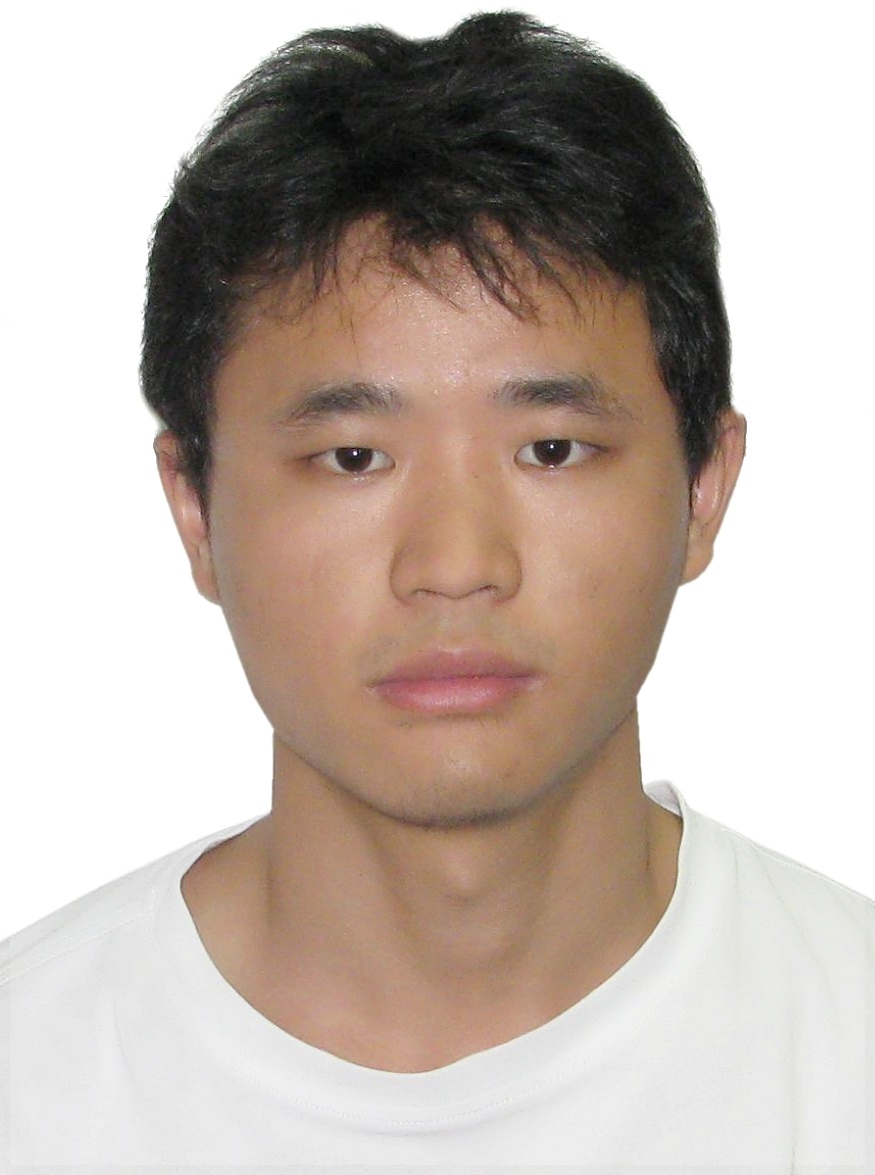}}]{Juyong Zhang}
is an associate professor in the School of Mathematical Sciences at University of Science and Technology of China. He received the BS degree from the University of Science and Technology of China in 2006, and the PhD degree from Nanyang Technological University, Singapore. His research interests include computer graphics, computer vision, and numerical optimization. He is an associate editor of The Visual Computer.
\end{IEEEbiography}
\begin{IEEEbiography}[{\includegraphics[width=1in]{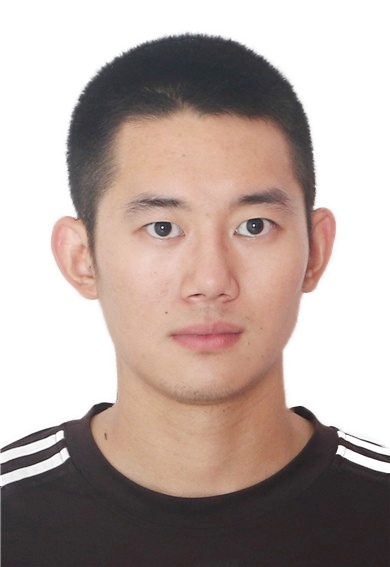}}]{Keyu Chen}
is a master student at the School of Mathematical Sciences, University of Science and Technology of China. Before that, he received his bachelor degree from the same University in 2018. His research interests include Computer Vision and Computer Graphics.
\end{IEEEbiography}
\begin{IEEEbiography}[{\includegraphics[width=1in]{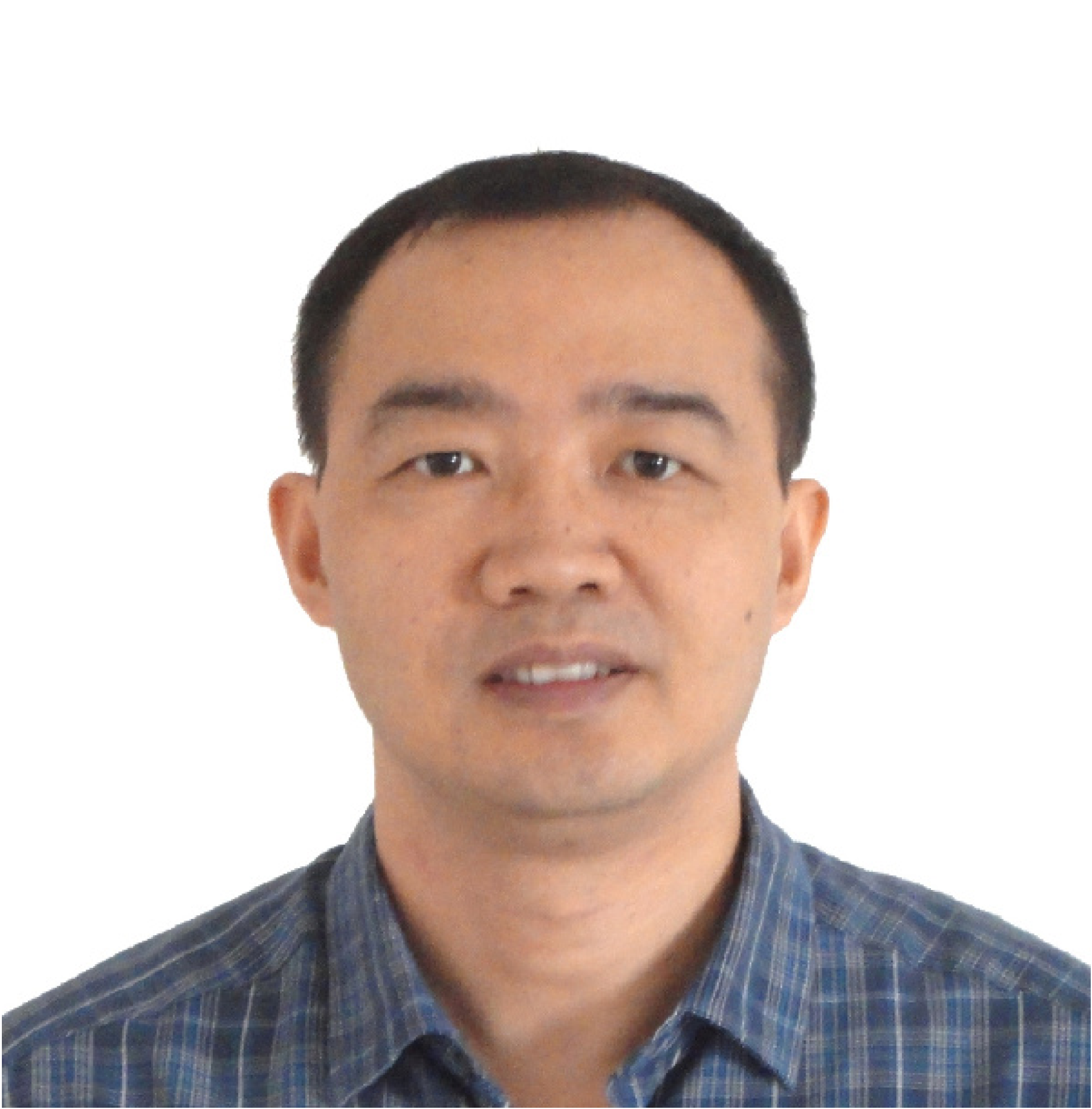}}]{Jianmin Zheng}
is an associate professor in the School of Computer Science and Engineering at Nanyang Technological University, Singapore. He received the BS and PhD degrees from Zhejiang University, China. His recent research focuses on T-spline technologies, digital geometric processing, reality computing, AR/VR, and AI
assisted part design for 3D printing. He is currently the programme director for the research
pillar of ML/AI under the HP-NTU Digital Manufacturing Corporate Lab. He is also a member of
executive committee of AsiaGraphics Association.
\end{IEEEbiography}

\end{document}